\documentclass[10pt,twocolumn,letterpaper]{article}

\usepackage{iccv}              
\usepackage{adjustbox}
\usepackage{multirow}
\usepackage{float}
\usepackage{makecell}
\usepackage[accsupp]{axessibility}
\usepackage{indentfirst}
\usepackage{comment}
\usepackage{multicol}

\definecolor{iccvblue}{rgb}{0.21,0.49,0.74}
\usepackage[pagebackref,breaklinks,colorlinks,allcolors=iccvblue]{hyperref}

\newcommand{\myparagraph}[1]{\vspace{3pt}\noindent\textbf{#1 ---}}

\def\paperID{14301} 
\def\confName{ICCV}
\def\confYear{2025}

\title{Controllable Latent Space Augmentation for Digital Pathology}
\author{%
  Sofiène Boutaj$^{1^{\diamond}}$, Marin Scalbert$^{2^*}$, Pierre Marza$^1$, Florent Couzinie-Devy$^3$,\\
  Maria Vakalopoulou$^1$, Stergios Christodoulidis$^1$\\
  \\
  $^1$MICS, CentraleSupélec – Université Paris-Saclay\\
  $^2$Bioptimus, Inc. $^3$VitaDX International \\ 
   \\ 
}

\begin{document}
\maketitle
\def\thefootnote{$\diamond$}\footnotetext{Corresponding email : \href{sofiene.boutaj@centralesupelec.fr}{sofiene.boutaj@centralesupelec.fr}}
\def\thefootnote{*}\footnotetext{Work done at CentraleSupélec and VitaDX.}

\begin{abstract}
\noindent
Whole slide image (WSI) analysis in digital pathology presents unique challenges due to the gigapixel resolution of WSIs and the scarcity of dense supervision signals. While Multiple Instance Learning (MIL) is a natural fit for slide-level tasks, training robust models requires large and diverse datasets. Even though image augmentation techniques could be utilized to increase data variability and reduce overfitting, implementing them effectively is not a trivial task. Traditional patch-level augmentation is prohibitively expensive due to the large number of patches extracted from each WSI, and existing feature-level augmentation methods lack control over transformation semantics. We introduce \textit{HistAug}, a fast and efficient generative model for controllable augmentations in the latent space for digital pathology. By conditioning on explicit patch-level transformations (e.g., hue, erosion), \textit{HistAug} generates realistic augmented embeddings while preserving initial semantic information. 
Our method allows the processing of a large number of patches in a single forward pass efficiently, while at the same time consistently improving MIL model performance. Experiments across multiple slide-level tasks and diverse organs show that \textit{HistAug} outperforms existing methods, particularly in low-data regimes. Ablation studies confirm the benefits of learned transformations over noise-based perturbations and highlight the importance of uniform WSI-wise augmentation. Code is available at \url{https://github.com/MICS-Lab/HistAug}.
\end{abstract}

\begin{figure}[t]
  \centering
  \includegraphics[width=\linewidth]{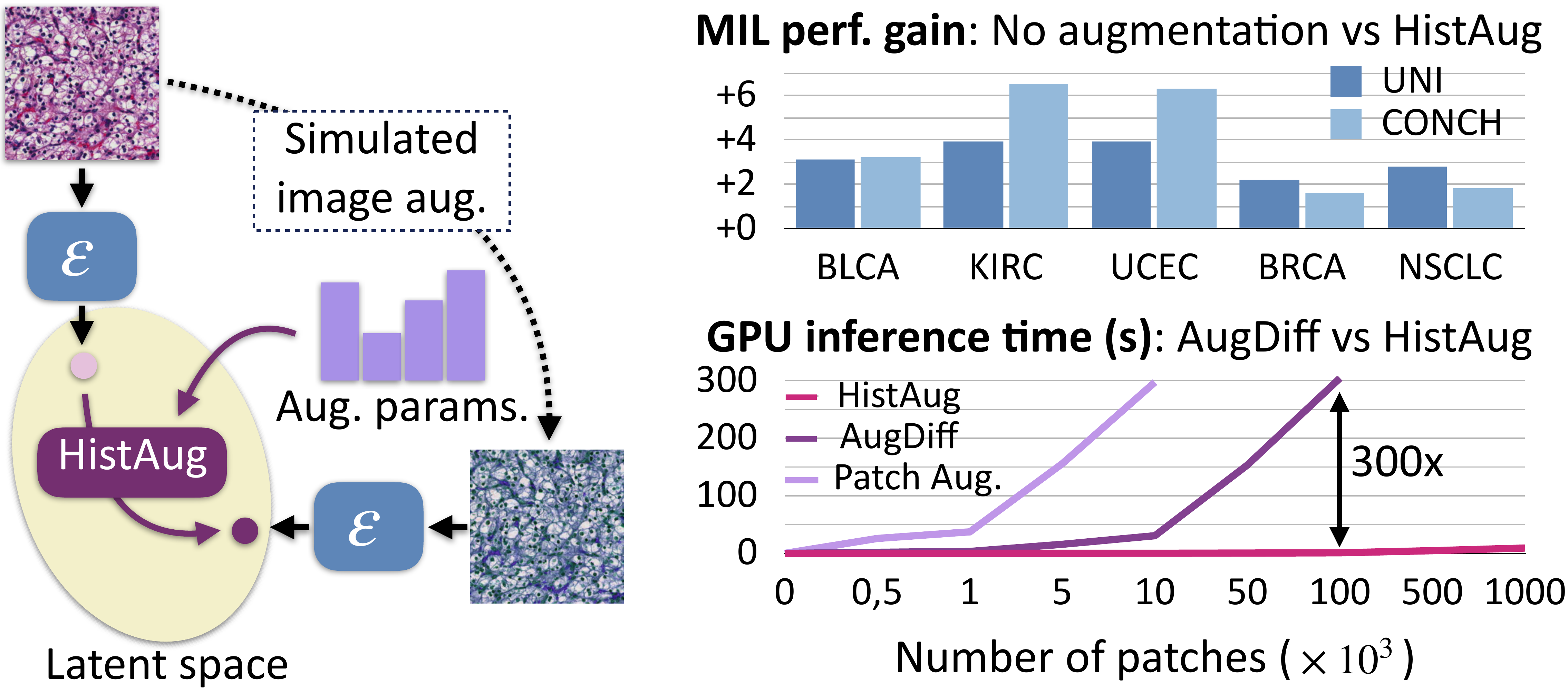}
  \caption{\textbf{HistAug}: We present a controllable and efficient method to augment WSI patches in the latent space of a vision encoder ($\mathcal{E}$) for histopathology MIL training.
  Such latent space augmentation leads to better downstream performance (gain in low-data shown here) as WSI datasets often lack data diversity. Our approach is faster and requires less memory than the state-of-the art diffusion model AugDiff~\cite{dai2024augdiff} and vanilla patch augmentation.}
  \label{fig:teaser_figure}
  \vspace{-5px}
\end{figure}

\vspace{-5px}
\section{Introduction}
\label{sec:intro}

In recent years, several deep learning methods have been proposed for a growing number of tasks in histopathology, including sub-typing, time-to-event endpoints, gene marker prediction, and others \cite{song2023artificial}. Most of these approaches leverage digitized histopathology tissue slides referred to as whole slide images (WSIs). These high-resolution, gigapixel-sized images provide details of the tissue micro-environment, enabling precise and comprehensive analysis. To efficiently process WSIs, a standard multi-step workflow has emerged: detect the tissue on the slide, divide it into patches, extract features from each patch, and finally perform a slide-level aggregation of the features and obtain a prediction from the aggregated features. These steps are often executed sequentially offline and the final model is trained on the feature space. On these grounds, many foundation models have been trained with self-supervised learning~\cite{chen2024towards, lu2024visual, vorontsov2024foundation, nechaev2024hibou, filiot2024phikon, zimmermann2024virchow2, alber2025novel} to serve as frozen patch-level feature extractors, removing the need for training task-specific vision encoders. Furthermore, since annotations are mainly available at the slide level and not for each patch individually, the aggregation models rely on Multiple Instance Learning (MIL) approaches~\cite{ilse2018attention, lu2021data, shao2021transmil}.
 
A primary challenge for deep learning models in digital pathology is the limited number of annotated slides for each specific task. Although patch-level augmentation can improve robustness and generalization, performing it \emph{online} for gigapixel WSIs is typically infeasible: it would require reading, transforming, and re-embedding on-the-fly tens or hundreds of thousands of patches per WSI. A common workaround is to pre-augment the patches offline and store multiple augmented versions of each WSI. However, this drastically increases both storage and preprocessing time, while offering limited diversity. Feature-level augmentation has thus emerged as a promising alternative, applying transformations directly to patch features in a way that simulates patch-level and pixel-level transformations (e.g. color jittering). The few existing solutions available are based on generative processes (\eg, diffusion-based ~\cite{dai2024augdiff} or GANs~\cite{zaffar2023embedding}) and suffer from two major limitations: (i) they lack direct control over transformations and (ii) they incur large memory and/or time overhead.

In this study, we propose a novel transformer-based model, named \textit{HistAug}, performing latent-space augmentation that can be used in MIL (Figure~\ref{fig:teaser_figure}). Given a patch embedding and a set of transformations together with their corresponding parameters (e.g. hue offsets), the model learns to predict the augmented patch embedding.
The precise control over transformation parameters offers flexibility in selecting which transformations to apply and adjusting their intensity based on specific tasks. The lightweight nature of our model allows a fast processing of multiple patch embeddings in parallel, making it easy to use within MIL training.

Our approach builds on the key insight that foundation models \emph{are not fully invariant} to image transformations~\cite{wölflein2024benchmarkingpathologyfeatureextractors, elphick2024latentrepresentationsfoundationmodels} and thus, that feature-space transforms can help during MIL training. This is validated on two state-of-the-art foundation models, the uni-modal vision encoder UNI~\cite{chen2024towards} and the vision-language model CONCH~\cite{lu2024visual}. Experiments on diverse histopathology datasets, with varying organs and tasks, show that \textit{HistAug} has a positive impact on downstream performance. Additionally, our model is much faster and less memory-intensive than the current state-of-the-art diffusion-based model AugDiff~\cite{dai2024augdiff} and naive patch augmentation. Since WSIs can also be processed at different magnification levels, we show that even if \textit{HistAug} generator is trained at a given magnification, it generalizes well to others. 

In summary, our main contributions are the following: \textit{(i)}
we introduce \emph{HistAug}, a novel generative model, designed to perform fast and controllable augmentation in the latent space. The model leverages a transformer architecture with cross-attention mechanisms to predict how patch-level transformations affect features of non-augmented patches, \textit{(ii)} the lightweight design of \textit{HistAug} allows to process $1M$ patches in less than $10$ seconds, making it easy to integrate into every MIL training setting, \textit{(iii)} the parameters of the augmentations applied to the features can be fully controlled allowing task-specific transformations without retraining the generator. At the same time, augmentations can be applied uniformly on a whole WSI, allowing, for example, consistent simulated color in the whole augmented bag.

A detailed validation of \textit{HistAug} on several MIL settings is performed using two foundation models, diverse histopathology datasets, and multiple magnifications. The quality of the simulated image augmentations is validated both quantitatively and through visualizations.

\section{Related Work}
\label{sec:related_work}

\myparagraph{Foundation models in histopathology} are trained with self-supervised learning on large patch-level datasets to later be used as general feature extractors. They can either be uni-modal vision encoders~\cite{chen2024towards, vorontsov2024foundation, nechaev2024hibou, filiot2024phikon, zimmermann2024virchow2, alber2025novel} or multi-modal vision-language models~\cite{lu2024visual, ding2024multimodal, zhou2024knowledge, ikezogwo2023quilt, huang2023visual, xiang2025vision, shaikovski2024prism}. UNI~\cite{chen2024towards} is a state-of-the-art VisionTransformer (ViT) model~\cite{vaswani2017attention, dosovitskiy2020image} trained with a DINOv2 objective~\cite{caron2021emerging, oquab2023dinov2} on a large dataset composed of over $100M$ patches extracted from more than $100,000$ WSIs representing $20$ tissue types. CONCH~\cite{lu2024visual} is a vision-language model trained on over $1M$ image-caption pairs with a contrastive objective~\cite{yu2022coca}. Its vision encoder is Transformer~\cite{vaswani2017attention}-based and can be leveraged along with the text encoder to perform many downstream tasks such as image captioning or text-to-image retrieval. Since our method is backbone-agnostic, in this work, we show its positive impact on the two different embedding spaces of UNI and CONCH as they are considered strong candidates to process WSIs. 

\myparagraph{Multiple Instance Learning (MIL)} is an efficient approach to train aggregator functions predicting information of interest from a bag of features. In histopathology in particular, such embeddings are extracted from WSI patches with foundation models as the ones presented above. MIL approaches involve simple attention mechanisms~\cite{ilse2018attention, lu2021data}, or even self-attention mechanisms through Transformer layers~\cite{shao2021transmil}. This MIL setting is convenient for training a model to efficiently aggregate patch-level features, but processing such a high number of embeddings simultaneously makes any image-based augmentation technique hard to consider due to compute and memory constraints.

We believe that image augmentations are too important to be ignored when considering the small scale of many WSI datasets, and thus, finding approaches to allow for more efficient training is important. To this end, we propose a modular and efficient method to augment patch features in the context of MIL training.

\myparagraph{Latent augmentations for MIL} have been proposed to mitigate the limitations expressed previously. By augmenting features directly in latent space to simulate real image augmentations, the benefits of data augmentations can be maintained while controlling computational costs. Feature-level augmentation methods can be divided into three families: (i) feature mixing, (ii) feature generation, and (iii) online representation sampling. Techniques in the first category rely on MixUp~\cite{zhang2017mixup} to interpolate features between instances \cite{chen2023rankmix}, instance prototypes \cite{yang2022remix, liu2024pseudo} or WSIs \cite{gadermayr2023mixupmil}, and thus cannot perform standard augmentations such as geometric transforms, color jittering or H\&E-tailored transforms, known to be useful in pathology~\cite{Tellez_2019}. Generative approaches are mainly based on GANs~\cite{zaffar2023embedding}, or diffusion models~\cite{dai2024augdiff}. Diffusion-based methods, such as AugDiff~\cite{dai2024augdiff} which is considered as the current state of the art, introduce controlled noise into feature embeddings and iteratively refine them. Their effectiveness at inference depends critically on the number of diffusion steps: too few results in minimal augmentation, while too much alters features. Additionally, diffusion models are computationally expensive in both time and memory, particularly in MIL settings where entire bags of features must be augmented, significantly limiting scalability. Both GANs and diffusion models lack explicit control over specific transformations, making them less effective for MIL tasks that require precise augmentation strategies. Recent self-supervised learning (SSL) approaches, such as SSRDL~\cite{tang2024self}, fall in the third category. SSRDL introduces an online representation sampling strategy to enhance MIL feature diversity. However, it requires training a dedicated patch encoder, making it incompatible with foundation models like UNI or CONCH, which offer strong generalization across pathology datasets.

To summarize, existing feature augmentation methods either lack transformation controllability, produce unrealistic augmentations, or require specialized feature extractors. Our work introduces a transformer-based generative model for feature augmentation that explicitly conditions augmentations on transformations while leveraging foundation models for feature extraction. Unlike previous methods, our approach enables consistent bag-wise augmentation while efficiently scaling to large datasets, providing a robust solution for MIL in histopathology.

\begin{figure*}
  \centering
  \includegraphics[width=1\linewidth]{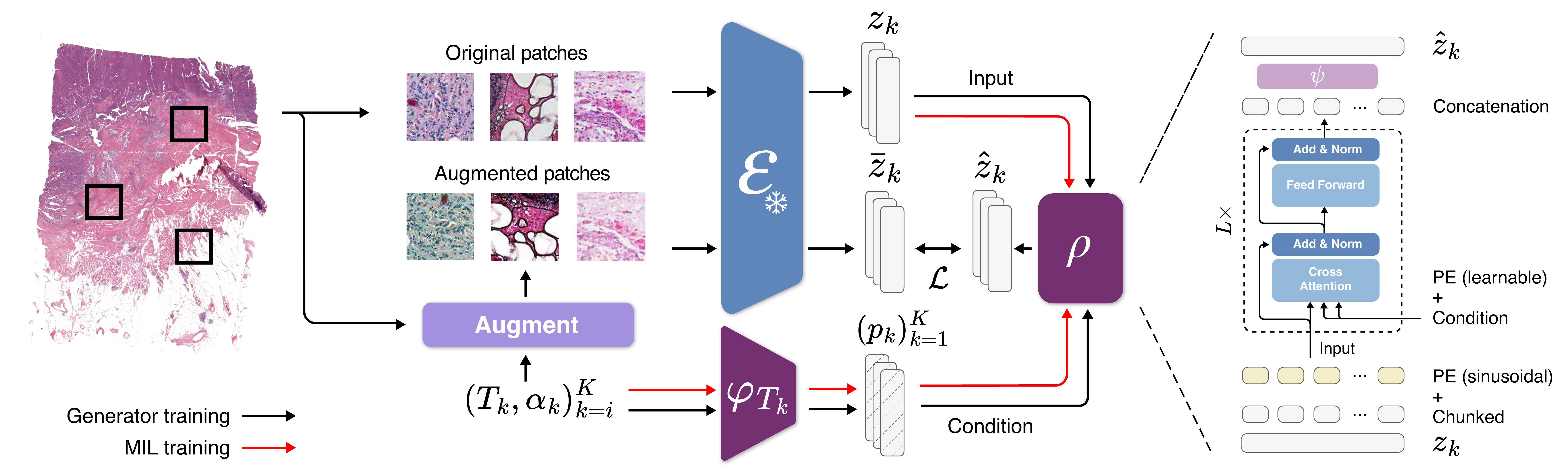}
  \caption{\textbf{HistAug Overview:} A frozen foundation model $\mathcal{E}$ (\eg, UNI) is used to encode both the original and augmented tissue patches to the latent spaces $z_k$ and $\tilde{z}_k$ respectively. The augmentation hyperparameters together with the original features are passed to a conditional transformer architecture that simulates the augmentation in the latent space producing $\hat{z}_k$. The whole pipeline is trained with a two-term loss function that combines a reconstruction and an identity loss. During MIL training, $\rho$ and the $ \varphi_{T_k}$ are frozen, and all operations are performed in the latent space.}
  \label{fig:method_figure}
  \vspace{-6px}
\end{figure*}

\section{Method}
Performing latent augmentations is particularly convenient in the context of MIL training where standard image-level augmentations can be time and memory-consuming. We propose to train a generator to transform latent features from a frozen foundation encoder into an augmented version. It is thus critical to learn to simulate standard image augmentations in latent space without distorting the information initially present in the image embedding. An overview of the method is presented in Figure~\ref{fig:method_figure}.

\subsection{Problem Setup}
Let us consider a set of $N$ available image augmentations $\{T_i\}_{i=1}^N$, and for each $T_i$, a space of available parameter values $A_{T_i}$ defining how the augmentation should be applied (e.g. the space of hue shifts when considering a hue transform). For transformations that only have a finite number of parameter values (e.g. flipping), $A_{T_i}$ will be a set of one-hot vectors.

Let $\mathbf{x} \in \mathbb{R}^{H \times W \times 3}$ be a patch from a WSI, and let $\mathcal{E}$ be a frozen feature extractor (i.e., foundation model), parametrized by weight $\theta_{\mathcal{E}}$, producing an embedding $\mathbf{z} {=} \mathcal{E}(\mathbf{x}; \theta_{\mathcal{E}}) \in \mathbb{R}^d$. 
We denote the function applying $K$ image transforms sequentially $\tau$ and define it as,
\begin{equation}
    \scalebox{0.9}{$
    \tau(\mathbf{x}; (T_k, \alpha_k)_{k=1}^K)
    \;{=}\;
    T_K\!\bigl(\cdots T_2\bigl(T_1(\mathbf{x}; \alpha_1);\alpha_2\bigr) \cdots; \alpha_K\bigr),
    $}
\end{equation}
where $(T_k, \alpha_k)_{k=1}^K$ is a sequence of transformations together with their associated parameter values, i.e. $T_k \sim \{T_i\}_{i=1}^N$ is a single transformation (e.g. change in hue)  with parameter $\alpha_k \sim A_{T_k}$ (e.g. hue shift value).

Our goal is to train a generator $\rho$ parametrized by weights $\theta_\rho$ to approximate the embedding of a transformed patch produced by the encoder $\mathcal{E}$:
\begin{equation}
    \rho\bigl(\mathbf{z},(T_k, \alpha_k)_{k=1}^K; \theta_\rho\bigr)
    \;\approx\;
    \mathcal{E}\bigl(\tau(\mathbf{x}; (T_k, \alpha_k)_{k=1}^K; \theta_{\mathcal{E}}\bigr).
\end{equation}

While simulating real image transformations in feature space, the generator should maintain as much of the initial encoded information as possible. As we will present later in this section, this will be enforced by training $\rho$ to reconstruct the identity, i.e. the input latent feature, if all parameters for the different transforms in the sequence have a specific value, i.e. if $(\alpha_k)_{k=1}^K {=} (\alpha_{\mathrm{id},k})_{k=1}^K$, where $\alpha_{\mathrm{id},k}$ is the parameter value for transformation $T_k$ leading to identity transform to be applied (i.e. the transform is not applied). We will then want the following,
\begin{equation}
    \rho\bigl(\mathbf{z}, (T_k, \alpha_{\mathrm{id},k})_{k=1}^K; \theta_\rho\bigr) \;\approx\; \mathbf{z}.
\end{equation}

\subsection{Generator Architecture}

\myparagraph{Chunked Inputs}
Since $\mathbf{z}$ is high-dimensional, we split it into $C$ segments,
\begin{equation}
    \mathbf{z} \;\mapsto\; (\mathbf{z}_i)_{i=1}^C,
    \quad 
    \mathbf{z}_i \;\in\; \mathbb{R}^{\tfrac{d}{C}}.
\end{equation}
Each chunk $\mathbf{z}_i$ is treated as a separate token in a transformer-based architecture. An ablation study on the interest of chunking is presented in Figure~\ref{fig:chunking_benefits} of the Supplementary Material. We encode the order of each token with sinusoidal positional encoding (PE)~\cite{vaswani2017attention}.

\myparagraph{Transformation Embeddings \& Cross-Attention}
Suppose we have a sequence of $K$ transformations $(T_k)_{k=1}^K$, each with parameters \(\alpha_k\). For each \(T_k\), we encode $\alpha_k$ into a parameter vector $\mathbf{p}_k \in \mathbb{R}^{\tfrac{d}{C}}$ with a linear projection layer $\varphi_{T_k}$ parametrized with weights $\theta_\varphi^{T_k}$. We have one projection layer per type of transformation: 

\begin{equation}
    \mathbf{p}_k = \varphi_{T_k}(\alpha_k; \theta_\varphi^{T_k}) = \theta_\varphi^{T_k} \cdot \alpha_k \in \mathbb{R}^{\tfrac{d}{C}},
\end{equation}
where $\alpha_k$ is a one-hot vector for transformations with a discrete number of parameters, and $\theta_\varphi^{T_k} \subset \theta_\rho$. To explicitly capture the order of transformations, we also add learnable positional embeddings to the $(\mathbf{p}_k)_{k=1}^K$.

The generator is composed of a sequence of $L$ transformer blocks $(\phi_j)_{i=j}^L$. Each $\phi_j$ block is parametrized by weights $\theta_\phi^j$ and performs cross-attention from the chunked tokens (queries) to the transformation tokens (keys/values), followed by a residual (skip) connection, layer normalization, and a feed-forward MLP with another skip connection,
\begin{equation}
\bigl(\Tilde{\mathbf{z}}_i^{j+1}\bigr)_{i=1}^C =
\phi_j \bigl((\Tilde{\mathbf{z}}_1^j)_{i=1}^C, (\mathbf{p}_k)_{k=1}^K; \theta_\phi^j \bigr),
\end{equation}
where $\Tilde{\mathbf{z}}_i^j$ is the representation token at block $j$ ($\Tilde{\mathbf{z}}_i^1$ = $\mathbf{z}_i$), and $\theta_\phi^j \subset \theta_\rho$. 
Such cross-attention mechanism allows sharing information between image features and augmentation parameter embeddings. After such $L$ blocks, we concatenate the updated chunked tokens back into $\mathbb{R}^d$, then apply an MLP head $\psi$ to obtain the final augmented feature as,
\begin{equation}
    \widehat{\mathbf{z}}
    = 
    \psi(\bigr[\Tilde{\mathbf{z}}_1^{L}, \dots, \Tilde{\mathbf{z}}_C^{L}\bigr ]; \theta_\psi),
\end{equation}
where $[ \cdot ]$ is the concatenation operator, and $\theta_\psi \subset \theta_\rho$.

\subsection{Objective Function}
The generator is trained to reconstruct the features produced by the frozen encoder $\mathcal{E}$ when given an augmented image, effectively mimicking the transformations we apply to the input. Additionally, it is trained to retain as much of the original feature information as possible, minimizing any distortion of the original latent representation. This is implemented as the combination of a \textbf{reconstruction loss} and an \textbf{identity loss}:
\begin{equation}
\begin{aligned}
\mathcal{L} 
&= \underbrace{\bigl\| \rho\bigl(\mathbf{z}, (T_k, \alpha_k)_{k=1}^K; \theta_\rho\bigr)
    \;-\; \mathcal{E}\bigl(\tau(\mathbf{x}; (T_k, \alpha_k)_{k=1}^K; \theta_{\mathcal{E}}\bigr)\bigr\|_2^2}_{\text{Reconstruction}} \\
&\quad + \lambda_{\mathrm{id}}\,
    \underbrace{\bigl\|\rho\bigl(\mathbf{z}, (T_k, \alpha_{\mathrm{id},k})_{k=1}^K;\theta_\rho\bigr) 
    \;-\; \mathbf{z}\bigr\|_2^2}_{\text{Identity}}.
\end{aligned}
\end{equation}

 The first term enforces that when the patch \(\mathbf{x}\) is transformed by \(T_1,\dots,T_K\) (with parameters \(\alpha_1,\dots,\alpha_K\)), the output from the generator aligns with the embedding from $\mathcal{E}$. The second term ensures that when no transformation is applied, the generator recovers the original embedding \(\mathbf{z}\).

\subsection{Integration into MIL Training}
Once $\rho$ is trained, we can augment any patch embedding in feature space, without recomputing image features. This allows us to efficiently augment data when training a model to aggregate patch features with MIL. Let $\{\mathbf{z}_m\}_{m=1}^M$ be a bag of $M$ embeddings extracted from a WSI. We distinguish two ways to augment patch features at training time:

\myparagraph{Instance-wise} Each patch is augmented with a different sequence of $K$ transformations sampled at random. For a given patch $\mathbf{z}_m$, we thus have a specific sequence of transformations $(T_k^m)_{k=1}^K$ and parameters $(\alpha_k^m)_{k=1}^K$, yielding
    $\widehat{\mathbf{z}}_m {=} \rho\bigl(\mathbf{z}_m, (T_k^m, \alpha_k^m)_{k=1}^K; \theta_\rho\bigr).$  
    
\myparagraph{Bag-wise (WSI-wise):} The same sequence of transformations $(T_k)_{k=1}^K$ and parameters $(\alpha_k)_{k=1}^K$ is sampled for all patches in the bag. We thus have $\widehat{\mathbf{z}}_m {=} \rho\bigl(\mathbf{z}_m, (T_k, \alpha_k)_{k=1}^K; \theta_\rho\bigr).$
    
While the \textit{Instance-Wise} mode increases data diversity within a bag, an advantage of augmenting \textit{WSI-Wise} is to preserve global consistency across the whole slide.

Finally, the augmented bag $\{\widehat{\mathbf{z}}_m\}_{m=1}^M$ can be fed into any MIL model (e.g., ABMIL \cite{ilse2018attention}, CLAM \cite{lu2021data}, DSMIL \cite{li2021dual}, TransMIL \cite{shao2021transmil}). Since the generator is lightweight, we can augment all patches from a WSI at training time in a single forward pass with minimal overhead.

\section{Experiments and Implementation Details}

\subsection{Datasets}
In this study, we use five TCGA datasets: BLCA (Bladder Urothelial Carcinoma), BRCA (Breast Invasive Carcinoma), NSCLC (Non-Small Cell Lung Carcinoma), UCEC (Uterine Corpus Endometrial Carcinoma) and KIRC (Kidney Renal Clear Cell Carcinoma). For training the generator, BLCA,  BRCA, and LUSC (Lung Squamous Cell Carcinoma) from NSCLC are used:  60\% of WSIs are sampled from each dataset, resulting in approximately 1200 training WSIs. Slides are processed at $10\times$ and $20\times$ magnification using the CLAM toolbox \cite{lu2021data}. We evaluated our method in two different downstream tasks: cancer subtyping (BRCA, NSCLC) and survival analysis (BLCA, UCEC, KIRC). 

\subsection{Generator Training}
We train two separate generators $\rho_{\text{UNI}}$ and $\rho_{\text{CONCH}}$ for the UNI \cite{chen2024towards} and CONCH~\cite{lu2024visual} feature spaces, respectively. In our experiments, AugDiff~\cite{dai2024augdiff} is trained on the same dataset.

\myparagraph{Transformations}
We apply a diverse set of transformations to image patches, extract their features using the respective feature extractor, and train the generator to reconstruct the augmented feature while being conditioned on both the original feature and transformation parameters. Each transformation is parameterized, and multiple transformations are applied sequentially in a randomized manner, ensuring a rich augmentation space. The generator learns to predict the feature embedding corresponding to the transformed patch while preserving the semantic structure of the input. The transformations we consider include: (i) \textbf{geometric transformations} such as rotations, horizontal and vertical flips, random cropping, as well as morphological operations such as dilation and erosion; (ii) \textbf{color transformations} that modify the brightness, contrast, hue, gamma, and saturation; (iii) \textbf{histology-specific transformation} specifically the HED transform \cite{pmlr-v143-faryna21a}, designed particularly for H\&E-stained images. Transformation details are shown in Table~\ref{tab:transforms_details} from the Supplementary Material.

\begin{table}[t]
\centering
\scriptsize
\caption{\textbf{Generator Evaluation}: Mean Cosine similarity between original and generated augmented features over $10k$ patches at 10X and 20X, alongside feature extractor invariance. The $\dagger$ symbol denotes out of training distribution. }
\label{tab:cosine_similarity}
\begin{adjustbox}{width=1\linewidth}
\begin{tabular}{ll|cc|cc}
\toprule
  &  & \multicolumn{2}{c|}{\textbf{Feature Reconstruction}} & \multicolumn{2}{c}{\textbf{Feature Invariance}} \\
\cmidrule(lr){3-4} \cmidrule(lr){5-6}
  &  & \textbf{10$\times$} & \textbf{20$\times$$\dagger$} & \textbf{10$\times$} & \textbf{20$\times$$\dagger$} \\
\midrule

\multirow{6}{*}{\rotatebox[origin=c]{90}{\textbf{UNI}}}
& BLCA & 81.0 & 74.9 & 11.9 & 11.6 \\
& BRCA  & 81.6 & 75.8 & 11.9 & 11.7 \\
& LUSC  & 81.1 & 74.9 & 12.6 & 11.9 \\
\cmidrule(lr){2-6}
& LUAD $\dagger$  & 80.3 & 74.6 & 15.0 & 11.3 \\
& UCEC $\dagger$  & 80.5 & 73.4 & 9.5 & 12.9 \\
& KIRC $\dagger$  & 80.5 & 72.1 & 13.0 & 11.0 \\
\midrule

\multirow{6}{*}{\rotatebox[origin=c]{90}{\textbf{CONCH}}}
& BLCA & 90.3 &88.1 & 20.3 & 27.7 \\
& BRCA  & 90.4 & 88.6 & 24.3 & 32.7 \\
& LUSC  & 90.4 & 87.0 & 19.3 & 27.7 \\
\cmidrule(lr){2-6}
& LUAD $\dagger$  & 90.2 & 87.6 & 24.9 & 23.7 \\
& UCEC $\dagger$  & 90.4 & 88.5 & 19.7 & 29.6 \\
& KIRC $\dagger$  & 89.9 & 87.5 & 23.1 & 29.8 \\
\bottomrule
\end{tabular}
\end{adjustbox}
\vspace{-5px}
\end{table}

\setlength\tabcolsep{2.5pt}
\begin{table}[t]
\centering
\scriptsize
\caption{ \textbf{Performance comparison across MIL models.} Mean and standard deviation of mean performance metrics aggregated across MIL models (Abmil, ClamMb, ClamSb, DSMIL, TransMIL). For BLCA, KIRC, and UCEC, C-index (\%) is used, while the AUC (\%) is used for the remaining datasets.}
\label{tab:performance_accross_models}
\begin{adjustbox}{width=1\linewidth}
\begin{tabular}{c l | c c c c c}
\toprule
\multicolumn{2}{c|}{} & \textbf{BLCA} & \textbf{KIRC} & \textbf{UCEC} & \textbf{BRCA} & \textbf{NSCLC} \\
\midrule
\multirow{11}{*}{\rotatebox[origin=c]{90}{\textbf{UNI}}} & \multicolumn{6}{l}{10\% Training} \\
\cmidrule(lr){2-7}
 & Base & 47.5$_{\pm1.9}$ & 58.5$_{\pm2.7}$ & 59.3$_{\pm4.7}$ & 86.1$_{\pm0.6}$ & 87.6$_{\pm3.8}$ \\
 & AugD & 49.9$_{\pm2.3}$ & \textbf{62.8$_{\pm4.7}$} & 61.9$_{\pm4.7}$ & 84.1$_{\pm5.5}$ & 86.8$_{\pm3.9}$ \\
 & PAug & 48.4$_{\pm1.4}$ & 60.1$_{\pm2.9}$ & 60.9$_{\pm5.5}$ & 88.2$_{\pm1.1}$ & 88.9$_{\pm4.3}$ \\
 & Ours (Inst) & 50.5$_{\pm1.0}$ & 61.3$_{\pm2.7}$ & 61.3$_{\pm5.9}$ & 88.2$_{\pm1.1}$ & 89.7$_{\pm3.1}$ \\
 & Ours (WSI) & \textbf{50.6$_{\pm1.6}$} & 62.5$_{\pm2.5}$ & \textbf{63.2$_{\pm3.6}$} & \textbf{88.3$_{\pm0.9}$} & \textbf{90.4$_{\pm3.6}$} \\
\cmidrule(lr){2-7}
 & \multicolumn{6}{l}{100\% Training} \\
\cmidrule(lr){2-7}
 & Base & 54.5$_{\pm3.7}$ & 65.8$_{\pm2.4}$ & 63.8$_{\pm4.2}$ & 92.7$_{\pm0.9}$ & \textbf{97.7$_{\pm1.1}$} \\
 & PAug & 56.7$_{\pm3.7}$ & 66.9$_{\pm1.5}$ & 64.7$_{\pm2.2}$ & \textbf{93.3$_{\pm0.3}$} & 93.5$_{\pm0.9}$ \\
 & Ours (Inst) & \textbf{60.3$_{\pm2.3}$} & 67.6$_{\pm1.5}$ & \textbf{65.5$_{\pm3.5}$} & 92.4$_{\pm0.2}$ & 97.3$_{\pm0.8}$ \\
 & Ours (WSI) & 59.5$_{\pm2.6}$ & \textbf{68.2$_{\pm1.1}$} & 64.8$_{\pm2.6}$ & 92.4$_{\pm0.3}$ & \textbf{97.7$_{\pm0.4}$} \\
\midrule
\multirow{11}{*}{\rotatebox[origin=c]{90}{\textbf{CONCH}}} & \multicolumn{6}{l}{10\% Training} \\
\cmidrule(lr){2-7}
 & Base & 50.8$_{\pm2.2}$ & 63.1$_{\pm3.0}$ & 58.6$_{\pm3.2}$ & 89.2$_{\pm3.5}$ & 92.8$_{\pm1.3}$ \\
 & AugD & 53.0$_{\pm1.7}$ & 65.9$_{\pm2.9}$ & 61.9$_{\pm0.9}$ & 90.1$_{\pm3.8}$ & 93.8$_{\pm0.5}$ \\
 & PAug & 54.2$_{\pm2.0}$ & 65.3$_{\pm4.0}$ & 64.5$_{\pm0.5}$ & 88.1$_{\pm3.0}$ & 93.2$_{\pm0.8}$ \\
 & Ours (Inst) & \textbf{54.5$_{\pm1.6}$} & 68.4$_{\pm3.1}$ & 63.7$_{\pm1.0}$ & 90.4$_{\pm1.2}$ & 93.9$_{\pm0.9}$ \\
 & Ours (WSI) & 54.1$_{\pm3.0}$ & \textbf{69.6$_{\pm3.0}$} & \textbf{64.9$_{\pm0.9}$} & \textbf{90.8$_{\pm1.7}$} & \textbf{94.6$_{\pm1.1}$} \\
\cmidrule(lr){2-7}
 & \multicolumn{6}{l}{100\% Training} \\
\cmidrule(lr){2-7}
 & Base & 58.0$_{\pm2.1}$ & 68.5$_{\pm1.5}$ & 60.1$_{\pm1.3}$ & 93.0$_{\pm1.7}$ & 97.9$_{\pm0.3}$ \\
 & PAug & 61.3$_{\pm1.5}$ & 69.9$_{\pm1.3}$ & 64.8$_{\pm5.0}$ & 92.4$_{\pm0.5}$ & 94.6$_{\pm1.1}$ \\
 & Ours (Inst) & \textbf{63.5$_{\pm1.6}$} & 70.1$_{\pm1.2}$ & 63.7$_{\pm2.9}$ & 93.5$_{\pm0.7}$ & \textbf{98.1$_{\pm0.3}$} \\
 & Ours (WSI) & 62.5$_{\pm0.3}$ & \textbf{71.2$_{\pm1.7}$} & \textbf{65.2$_{\pm1.7}$} & \textbf{93.7$_{\pm0.5}$} & 97.8$_{\pm0.6}$ \\
\bottomrule
\end{tabular}
\end{adjustbox}
\vspace{-2px}
\end{table}
\subsection{MIL Training}
We train MIL models using both UNI \cite{chen2024towards} and CONCH \cite{lu2024visual} foundation models as vision encoders. We generate five different data folds for each dataset and evaluate five MIL architectures: Abmil \cite{ilse2018attention}, CLAM \cite{lu2021data}, DSMIL \cite{li2021dual}, and TransMIL \cite{shao2021transmil}, each leveraging different aggregation mechanisms to predict slide-level outcomes. For each MIL model, we compare six augmentation settings: \textbf{Base} refers to models trained without augmentation. \textbf{AugDiff (AugD)} applies feature augmentation using the state-of-the-art AugDiff method \cite{dai2024augdiff}. \textbf{Patch augmentation (PAug)} applies transformations in image space offline, and stores the augmented features before MIL training. HistAug (Ours) includes both \textbf{instance-wise (Inst)} and \textbf{WSI-wise (WSI)} feature augmentation, leveraging our trained generator. The last baseline performs \textbf{noise-based perturbations (Noise)}. 

\subsection{Evaluation Protocol}
\myparagraph{Generator Evaluation} We evaluate the performance of the generator using two key metrics. \textbf{Feature reconstruction} where we randomly sample 10,000 patches from WSIs that were not used during training and we compute the cosine similarity between the generated augmented feature $\widehat{z}$ and the original augmented feature $\bar{\mathbf{z}}$. A high cosine similarity indicates that the generator accurately replicates the effect of image-space transformations in feature space. \textbf{Feature extractor invariance} where we compute the cosine similarity between the original embedding $\mathbf{z} =\mathcal{E}(\mathbf{x}; \theta_\varepsilon)$ and the real augmented feature $\bar{\mathbf{z}}$. Lower similarity values suggest that the feature extractor is sensitive to transformations in its latent space~\cite{wölflein2024benchmarkingpathologyfeatureextractors, elphick2024latentrepresentationsfoundationmodels}. 

\myparagraph{Downstream Tasks} For each model, we conduct training across five different folds. Area Under the Curve (AUC) and C-Index are reported for subtyping and survival analysis respectively. Moreover, we perform experiments on two different data regimes namely (i)  \textbf{10\% data regime} at which the models are trained using only 10\% of the available labeled training samples, while the testing sets remain the same and (ii) \textbf{100\% data regime} where the models are trained on the full set of available training data.

\section{Results}

\begin{table*}[t]
\scriptsize
\centering
\caption{\textbf{Performance comparison across datasets}. Mean and standard deviation of mean performance metrics aggregated across tasks (BLCA, KIRC, UCEC for survival; BRCA, NSCLC for classification) are reported for each MIL model and feature augmentation strategy.}
\label{tab:performance_accross_tasks}
\begin{adjustbox}{width=0.9\textwidth}
\begin{tabular}{l|ccccc|ccccc}
\toprule
 & \multicolumn{5}{c|}{\textbf{UNI}} & \multicolumn{5}{c}{\textbf{CONCH}} \\
\cmidrule(lr){2-6}\cmidrule(lr){7-11}
Augmentation & Abmil & ClamMb & ClamSb & DSMIL & TransMIL& Abmil & ClamMb & ClamSb & DSMIL & TransMIL \\
\midrule
\multicolumn{11}{l}{Survival (C-Index, \%) -- 10\% Training} \\
\midrule
Base & $52.5_{\pm4.6}$ & $55.2_{\pm4.3}$ & $51.8_{\pm4.3}$ & $57.5_{\pm8.0}$ & $58.5_{\pm6.6}$ & $55.4_{\pm4.4}$ & $56.9_{\pm4.3}$ & $57.4_{\pm7.1}$ & $58.8_{\pm7.2}$ & $59.0_{\pm4.2}$ \\
AugD & $\mathbf{58.3}_{\pm3.7}$ & $\mathbf{59.4}_{\pm8.0}$ & $56.3_{\pm5.0}$ & $60.5_{\pm9.7}$ & $56.5_{\pm6.7}$ & $59.8_{\pm5.7}$ & $60.1_{\pm6.1}$ & $59.7_{\pm6.5}$ & $62.6_{\pm6.0}$ & $59.3_{\pm3.1}$ \\
PAug & $53.9_{\pm4.1}$ & $54.9_{\pm5.0}$ & $54.0_{\pm3.9}$ & $58.8_{\pm9.0}$ & $\mathbf{60.7}_{\pm7.3}$ & $58.9_{\pm5.7}$ & $62.5_{\pm5.4}$ & $61.4_{\pm5.5}$ & $64.3_{\pm5.8}$ & $59.5_{\pm4.3}$ \\
Ours (Inst) & $56.0_{\pm5.1}$ & $59.1_{\pm6.0}$ & $53.7_{\pm3.3}$ & $59.7_{\pm6.6}$ & $59.9_{\pm7.4}$ & $60.3_{\pm5.7}$ & $63.7_{\pm5.5}$ & $\mathbf{62.9}_{\pm6.9}$ & $63.3_{\pm6.6}$ & $60.9_{\pm4.9}$ \\
Ours (WSI) & $56.7_{\pm6.0}$ & $58.2_{\pm6.2}$ & $\mathbf{57.9}_{\pm3.6}$ & $\mathbf{60.7}_{\pm7.0}$ & $60.4_{\pm7.4}$ & $\mathbf{61.9}_{\pm6.9}$ & $\mathbf{64.5}_{\pm5.4}$ & $61.7_{\pm9.2}$ & $\mathbf{64.8}_{\pm6.4}$ & $\mathbf{61.3}_{\pm5.3}$ \\
\midrule
\multicolumn{11}{l}{Survival (C-Index, \%) -- 100\% Training} \\
\midrule
Base & $60.6_{\pm6.5}$ & $62.5_{\pm3.2}$ & $60.4_{\pm6.7}$ & $59.5_{\pm7.6}$ & $63.7_{\pm4.2}$ & $60.8_{\pm5.1}$ & $63.6_{\pm4.7}$ & $62.3_{\pm3.0}$ & $60.5_{\pm5.3}$ & $63.8_{\pm4.9}$ \\
PAug & $\mathbf{63.6}_{\pm3.8}$ & $61.9_{\pm5.4}$ & $63.1_{\pm3.1}$ & $59.9_{\pm6.9}$ & $\mathbf{65.5}_{\pm3.6}$ & $65.8_{\pm3.6}$ & $\mathbf{68.7}_{\pm4.2}$ & $\mathbf{66.4}_{\pm3.5}$ & $64.0_{\pm4.1}$ & $61.7_{\pm4.8}$ \\
Ours (Inst) & $63.1_{\pm4.0}$ & $61.9_{\pm4.3}$ & $\mathbf{66.9}_{\pm2.9}$ & $\mathbf{66.2}_{\pm4.1}$ & $64.3_{\pm1.3}$ & $65.2_{\pm3.9}$ & $67.8_{\pm3.1}$ & $66.0_{\pm3.1}$ & $\mathbf{66.4}_{\pm3.0}$ & $63.4_{\pm3.9}$ \\
Ours (WSI) & $63.1_{\pm4.8}$ & $\mathbf{62.7}_{\pm5.1}$ & $65.0_{\pm2.0}$ & $65.4_{\pm4.9}$ & $64.7_{\pm2.7}$ & $\mathbf{67.3}_{\pm3.6}$ & $67.6_{\pm4.3}$ & $65.5_{\pm3.8}$ & $66.3_{\pm4.3}$ & $\mathbf{65.0}_{\pm2.3}$ \\
\midrule
\multicolumn{11}{l}{Classification (AUC, \%) -- 10\% Training} \\
\midrule
Base & $87.5_{\pm1.8}$ & $89.0_{\pm2.0}$ & $88.8_{\pm2.4}$ & $83.2_{\pm2.0}$ & $85.8_{\pm0.6}$ & $\mathbf{93.4}_{\pm1.8}$ & $91.7_{\pm1.4}$ & $91.3_{\pm1.1}$ & $86.8_{\pm4.5}$ & $91.7_{\pm0.3}$ \\
AugD & $88.2_{\pm0.9}$ & $87.3_{\pm0.1}$ & $88.2_{\pm1.9}$ & $76.2_{\pm3.0}$ & $\mathbf{87.3}_{\pm0.7}$ & $92.8_{\pm0.9}$ & $93.2_{\pm0.6}$ & $93.5_{\pm1.1}$ & $88.1_{\pm5.6}$ & $92.1_{\pm0.9}$ \\
PAug & $89.3_{\pm2.4}$ & $91.1_{\pm1.6}$ & $91.2_{\pm1.7}$ & $85.7_{\pm1.4}$ & $85.4_{\pm2.4}$ & $91.2_{\pm3.0}$ & $92.1_{\pm1.3}$ & $91.1_{\pm3.0}$ & $87.6_{\pm5.1}$ & $91.5_{\pm0.5}$ \\
Ours (Inst) & $90.3_{\pm0.9}$ & $91.0_{\pm1.6}$ & $90.3_{\pm2.1}$ & $\mathbf{87.1}_{\pm0.1}$ & $85.9_{\pm1.1}$ & $92.7_{\pm1.8}$ & $93.0_{\pm1.8}$ & $92.6_{\pm1.9}$ & $90.5_{\pm2.5}$ & $92.0_{\pm0.8}$ \\
Ours (WSI) & $\mathbf{90.8}_{\pm1.6}$ & $\mathbf{91.3}_{\pm2.4}$ & $\mathbf{91.4}_{\pm2.4}$ & $86.3_{\pm0.7}$ & $87.1_{\pm0.4}$ & $93.3_{\pm1.6}$ & $\mathbf{93.6}_{\pm2.1}$ & $\mathbf{93.5}_{\pm2.1}$ & $\mathbf{90.6}_{\pm3.2}$ & $\mathbf{92.6}_{\pm0.4}$ \\
\midrule
\multicolumn{11}{l}{Classification (AUC, \%) -- 100\% Training} \\
\midrule
Base & $\mathbf{95.4}_{\pm2.4}$ & $\mathbf{95.8}_{\pm2.8}$ & $\mathbf{96.0}_{\pm2.5}$ & $94.5_{\pm3.6}$ & $94.2_{\pm1.3}$ & $95.9_{\pm2.1}$ & $\mathbf{96.3}_{\pm2.1}$ & $95.8_{\pm1.7}$ & $94.0_{\pm4.2}$ & $\mathbf{95.3}_{\pm2.3}$ \\
PAug & $93.7_{\pm0.1}$ & $93.5_{\pm0.5}$ & $93.8_{\pm0.6}$ & $92.7_{\pm0.8}$ & $93.5_{\pm0.1}$ & $93.8_{\pm1.0}$ & $94.1_{\pm1.2}$ & $93.8_{\pm1.2}$ & $93.9_{\pm1.4}$ & $91.9_{\pm0.6}$ \\
Ours (Inst) & $95.0_{\pm2.9}$ & $94.8_{\pm2.3}$ & $95.0_{\pm2.6}$ & $95.3_{\pm2.6}$ & $94.1_{\pm1.7}$ & $95.9_{\pm2.5}$ & $96.0_{\pm2.5}$ & $95.8_{\pm1.9}$ & $\mathbf{96.3}_{\pm1.9}$ & $95.0_{\pm2.8}$ \\
Ours (WSI) & $94.8_{\pm2.7}$ & $95.0_{\pm2.5}$ & $94.9_{\pm2.4}$ & $\mathbf{95.4}_{\pm3.0}$ & $\mathbf{95.3}_{\pm2.4}$ & $\mathbf{96.2}_{\pm2.2}$ & $96.1_{\pm2.4}$ & $\mathbf{96.1}_{\pm1.9}$ & $95.7_{\pm1.8}$ & $94.8_{\pm2.0}$ \\
\bottomrule
\end{tabular}
\end{adjustbox}
\end{table*}
\subsection{Generator Evaluation}

Table~\ref{tab:cosine_similarity} reports our performance in reconstructing augmented features at 10$\times$ and 20$\times$ magnification. Results demonstrate high cosine similarity between original and generated augmented features, consistently exceeding $80$ for UNI and approximately $90$ for CONCH. This is achieved despite the significant impact of the transformations on the latent space of feature extractors, i.e. cosine similarity drops to approximately $11$ to $15$ for UNI and around $20$ for CONCH, highlighting the sensitivity of these foundation models to sequences of transformations. This implies that, despite the non-invariance of the UNI and CONCH feature spaces, our generator can successfully capture and reconstruct these altered representations.

Importantly, although our generator was trained using patches extracted at 10$\times$ magnification, it can also reconstruct features from images at a higher magnification (i.e., 20$\times$). As seen in Table~\ref{tab:cosine_similarity}, the cosine similarity at 20$\times$ remains high (approximately $75$ for UNI and $88$ for CONCH), showcasing strong cross-scale generalization capabilities. 

Furthermore, the effectiveness of the generator extends to external datasets on different organs. For instance, results on external lung (LUAD), kidney (KIRC), and endometrial (UCEC) datasets demonstrate similarly high cosine similarity (around $80$ for UNI and $90$ for CONCH at 10$\times$), underscoring the robustness and generalizability of our generator to diverse histopathology datasets and tissue types unseen at training time. Tables~\ref{tab:cosine_similarity_10X} \&~\ref{tab:cosine_similarity_20X} in the Supplementary Material provide cosine similarity bootstrap confidence intervals.

\begin{figure}
    \centering
    \includegraphics[width=1\linewidth]{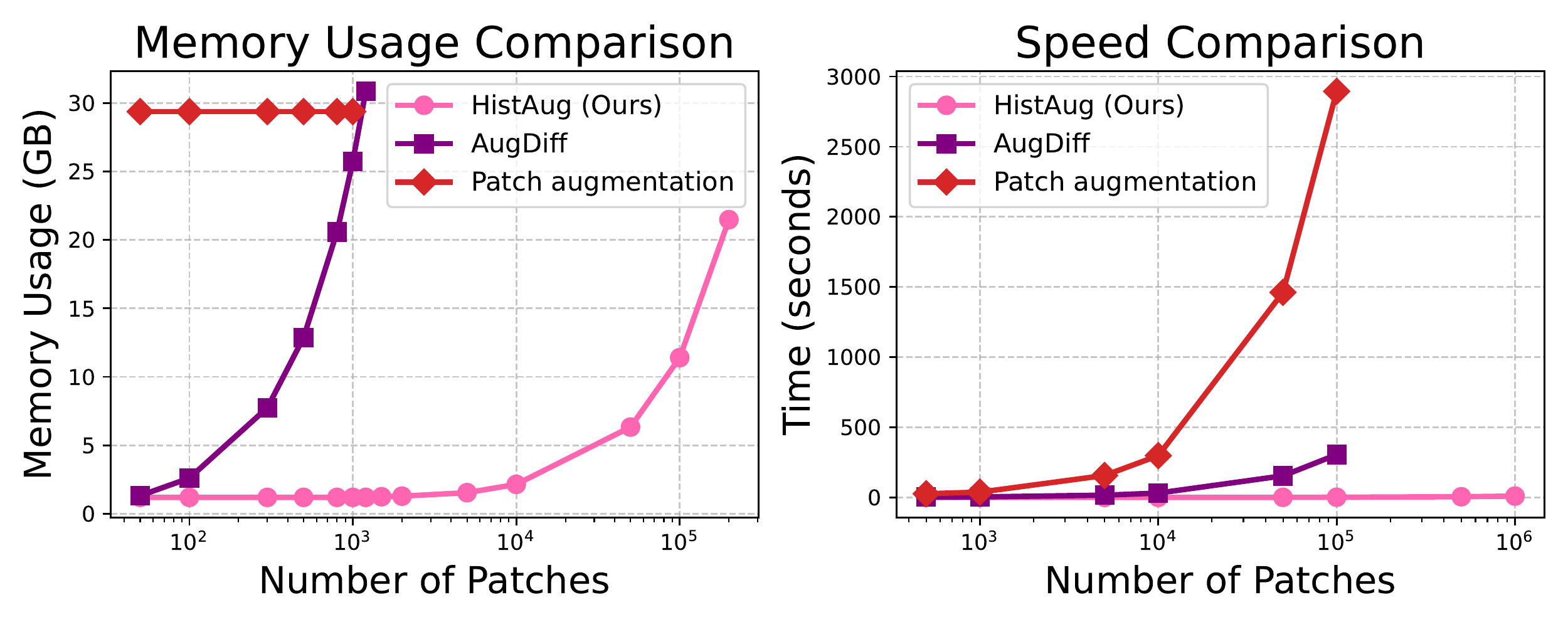}
    \caption{\textbf{Comparison in forward pass efficiency}: HistAug is significantly faster than AugDiff~\cite{dai2024augdiff} and requires less GPU memory to augment a set of features in a single parallelized forward pass. The experiments were conducted on a V100 32GB GPU}
    \label{fig:efficiency}
    \vspace{-3px}
\end{figure}

\subsection{MIL Evaluation}

Tables~\ref{tab:performance_accross_models} and \ref{tab:performance_accross_tasks} report MIL performance averaged across MIL models and datasets. At 10\% training data, instance-wise and WSI-wise augmentations with our method consistently outperform the baseline, AugDiff and PAug across nearly all cancer types and vision encoders. For instance, WSI-wise augmentation improves survival prediction (C-index) notably, from $59.3$ to $63.2$ for UNI and from $58.6$ to $64.9$  for CONCH in UCEC. Similar performance improvements are observed in classification tasks at $10\%$ training data, with WSI-wise augmentation achieving gains such as an increase from $86.1$ to $88.3$ for UNI in BRCA and from $92.8$ to $94.6$ for CONCH in NSCLC. Moreover, HistAug matches or outperforms PAug as it introduces more diverse augmentations on-the-fly during training, unlike PAug which relies on precomputed and stored features.

At 100\% training data, our method also yields improvements, particularly for survival prediction tasks e.g., instance-wise augmentation improves UNI-based BLCA from $54.5$ to $59.5$. For classification tasks, improvements are smaller, likely due to the already high baseline performance leaving limited room for further gains. AugDiff is not evaluated at 100\% training data due to its high computational cost (at $10\times$ magnification it requires over $60$ GPU hours per fold for BLCA dataset, see Section~\ref{subsec:speed_comparison}). An extensive experiment with AugDiff with 100\% of the data can be found in the Supplementary Material in Table~\ref{tab:augdiff_scaling}. Additional results on the BRACS dataset \cite{bracs_dataset} can be found in the Supplementary Material in Table~\ref{tab:bracs_res}.

Finally, we compare our method with SSRDL~\cite{tang2024self}, a state-of-the-art approach for Online Representation Sampling (ORS), on the TCGA-EGFR dataset (Table~\ref{tab:tcga_egfr}), as it is the only one with data splits provided in~\cite{tang2024self}. A crucial advantage of \textit{HistAug} is that, unlike SSRDL, it does not require to train a patch encoder along with the augmentation model. Instead, we can leverage powerful, histopathology-specific foundation models such as UNI to train a strong baseline already outperforming SSRDL, which is then improved with our augmented features. Tables~\ref{tab:10percent_10X} \&~\ref{tab:100percent_10X} in the Supplementary Material provide detailed results.

\subsection{Speed and Memory Comparison}\label{subsec:speed_comparison}
Since latent augmentation is performed at every training step in MIL, computational efficiency is a key consideration. For this reason, we compare the time and memory requirements of \textit{HistAug}, AugDiff, and naive patch augmentation — in which transformations are applied in the pixel space and features are then extracted by a foundation model — for augmenting a bag of features on a single $32$GB V100 GPU. As shown in Figure~\ref{fig:efficiency}, \textit{HistAug} is significantly faster than AugDiff while using much less GPU memory. \textit{HistAug} can process up to $1$M input patches in less than $10$ seconds, while in the same time, AugDiff can only augment fewer than $5$k patches. When considering a realistic number of patches as required in MIL training, i.e. $50$k--$100$k patches, \textit{HistAug} is approximately $300\times$ faster than AugDiff. Moreover, AugDiff reaches memory saturation on a $32$GB GPU when augmenting just $1$k patches in parallel, while \textit{HistAug} can handle batches of up to $200$k patches before reaching the same limit. Furthermore, patch augmentation scales poorly, with high GPU memory usage ($\sim32$\,GB) and long runtimes (3k\,s for 100k patches), making it impractical for MIL due to repeated foundation model forward passes. In addition to leading to better MIL performance, \textit{HistAug} is thus more convenient to use at training time.

\begin{table}
    \centering
    \scriptsize    \caption{\textbf{Comparison with SSRDL} on the TCGA-EGFR dataset. Performance is measured using ROC AUC.}
    \label{tab:tcga_egfr}
    \begin{tabular}{lccc}
        \toprule
        \textbf{Model} & \textbf{SSDRL} & \textbf{Baseline (UNI)} & \textbf{Ours (UNI)} \\
        \midrule
        TransMIL &  79.7  & 86.5 & \textbf{87.9} \\
        CLAM     & 83.1 & 86.5  & \textbf{89.4} \\
        \bottomrule
    \end{tabular}

\end{table}
\subsection{Underlying properties of latent augmentations}
\myparagraph{Cross-magnification Generalization} Our generator trained at 10$\times$ magnification also improves MIL performance when applied at 20$\times$ without additional training (Table \ref{tab:20X}). For example, WSI-wise augmentation significantly improves the C-index in UCEC from $60.5$ to $67.1$ (CONCH embeddings) in this out-of-distribution setting.

\myparagraph{Superiority to Noise Perturbation} To verify that our method is not simply altering features without meaningful structure, we introduced a noise perturbation baseline which adds random noise to features. As shown in Table~\ref{tab:noise_transposed}, our method (Inst and WSI) consistently outperfoms it across both 10\% and 100\% training data settings. Noise perturbation does not significantly improve performance, showing that random perturbations alone are insufficient. This further underscores the effectiveness and necessity of our learned augmentation strategies. Tables~\ref{tab:10percent_10X} and~\ref{tab:100percent_10X} in the Supplementary Material present full results.

\setlength\tabcolsep{2.0pt}
\begin{table}[t]
\centering
\caption{\textbf{Gains at the 20$\times$ OOD magnification} from augmented features highligh the generalizability properties of \textit{HistAug}.}
\label{tab:20X}
{ \footnotesize
    \begin{tabular}{l|cc|cc|cc|cc|cc}
    \toprule
      & \multicolumn{6}{c|}{\textbf{Survival (C-Index)}} & \multicolumn{4}{c}{\textbf{Classification (AUC)}} \\
    \cmidrule(lr){2-7}\cmidrule(lr){8-11}
      & \multicolumn{2}{c|}{BLCA} & \multicolumn{2}{c|}{KIRC} & \multicolumn{2}{c|}{UCEC} & \multicolumn{2}{c|}{BRCA} & \multicolumn{2}{c}{NSCLC} \\
    \cmidrule(lr){2-3}\cmidrule(lr){4-5}\cmidrule(lr){6-7}\cmidrule(lr){8-9}\cmidrule(lr){10-11}
     Model & 10\% & 100\% & 10\% & 100\% & 10\% & 100\% & 10\% & 100\% & 10\% & 100\% \\
    \midrule
    {\scriptsize\textbf{UNI}}
     & +$2.9$ & +$4.3$ & +$3.6$ & +$2.6$ & +$3.6$ & +$3.5$ & +$1.7$ & +$0.1$ & +$0.8$ & +$0.8$ \\
    \midrule
    {\scriptsize\textbf{CONCH}}
     & +$4.5$ & +$8.2$ & +$4.3$ & +$2.8$ & +$3.0$ & +$6.6$ & +$0.5$ & +$0.5$ & +$2.6$ & +$0.0$ \\
    \bottomrule
    \end{tabular}
}
\end{table}

\begin{table}
\centering
\scriptsize
\caption{\textbf{Gain over Noise} for UNI and CONCH at 10\% vs.\ 100\% training at $10 \times$ magnification for the two \textit{HistAug} variants.}
\label{tab:noise_transposed}
\begin{adjustbox}{width=1\linewidth}
    
\begin{tabular}{l|cc|cc|cc|cc}
\toprule
 & \multicolumn{2}{c}{\textbf{10\% UNI}} & \multicolumn{2}{c}{\textbf{10\% CONCH}} 
 & \multicolumn{2}{c}{\textbf{100\% UNI}} & \multicolumn{2}{c}{\textbf{100\% CONCH}} \\
\cmidrule(lr){2-3} \cmidrule(lr){4-5} \cmidrule(lr){6-7} \cmidrule(lr){8-9}
 & Inst & WSI & Inst & WSI & Inst & WSI & Inst & WSI \\
\midrule
BLCA  
 & +3.0 & +3.1 & +2.8 & +2.3 & +4.3 & +3.5 & +4.3 & +3.3 \\
KIRC  
 & +3.1 & +4.3 & +5.2 & +6.4 & +1.7 & +2.3 & +0.5 & +1.5 \\
UCEC  
 & +2.3 & +4.2 & +4.9 & +6.1 & +2.6 & +1.9 & +2.7 & +4.2 \\
BRCA  
 & +1.6 & +1.7 & -0.4 & +0.3 & -0.5 & -0.5 & +0.1 & +0.3 \\
NSCLC  
 & +0.7 & +1.4 & +1.3 & +2.0 & -0.2 & +0.2 & +1.1 & +0.8 \\
\bottomrule
\end{tabular}
\end{adjustbox}
\end{table}

\subsection{Visualizing the learned latent transformations}
\begin{figure}[t]
  \centering
  \includegraphics[width=1\linewidth]{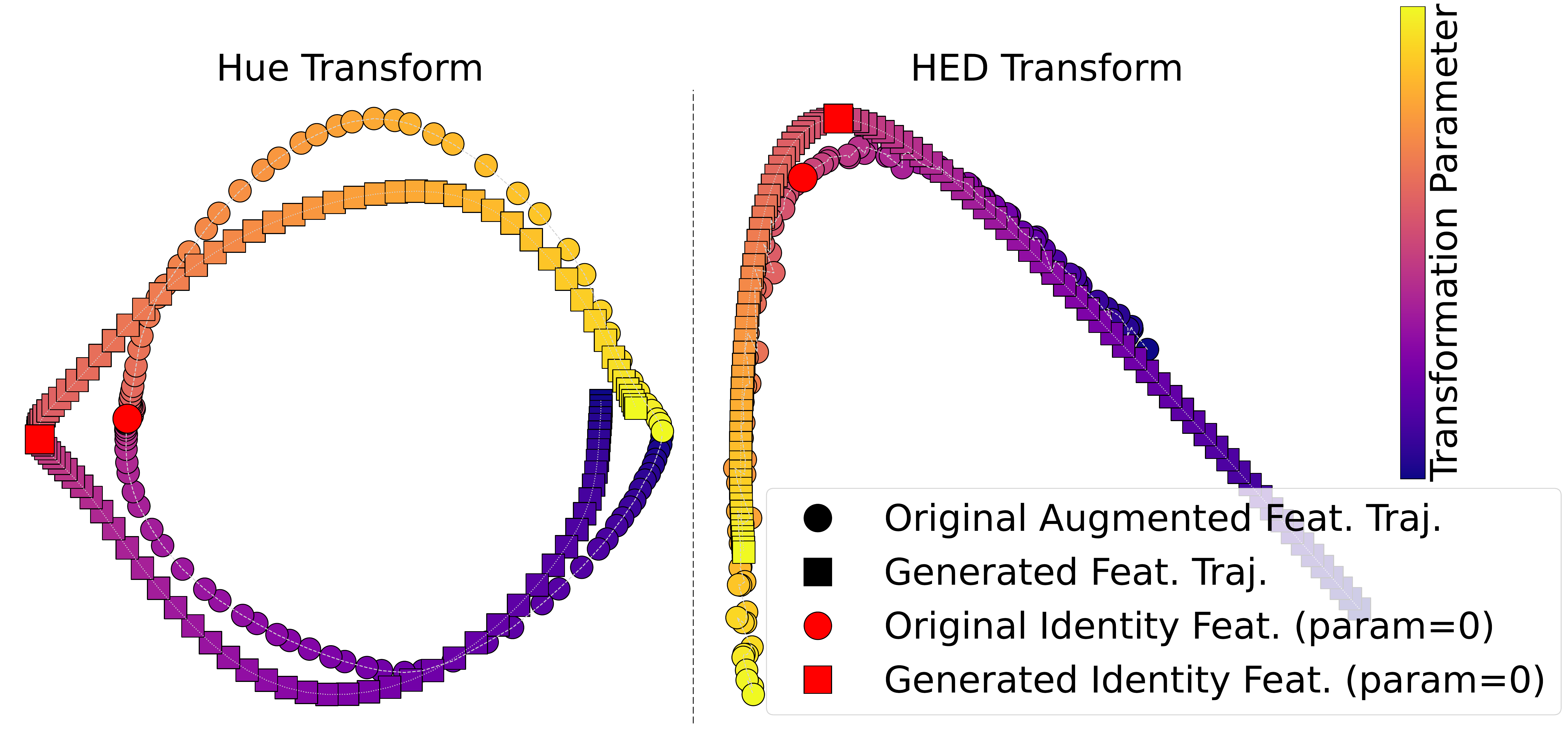}
  \caption{\textbf{Trajectories in augmentation space}: \textbf{Left} --  PCA trajectories of Hue transform for UNI. \textbf{Right} -- PCA trajectories of HED transform for CONCH. Circles represent the original augmented features extracted from the foundation model, while squares denote the features generated by our augmentation model. 
  }
  \label{fig:pca_trajectories}
\end{figure}

\myparagraph{Trajectories in augmentation space} An important characteristic of \textit{HistAug} is controllability. We visualize how augmented features evolve when navigating the space of parameter values. For a given patch and transformation, we sample a set of parameter values and, for each of them, both apply the original transformation followed by feature extraction with the encoder, and generate the augmented features with \textit{HistAug}. We apply a $2$-dim PCA on the whole set of original and generated augmented features and visualize the trajectory of latent codes. Trajectories in Figure~\ref{fig:pca_trajectories} are close between original and generated features, highlighting the controllability of \textit{HistAug}. Additional visualizations are shown in Figure~\ref{fig:pca_trajectories_supp_mat} in the Supplementary Material.

\begin{figure}
    \centering
    \includegraphics[width=1\linewidth]{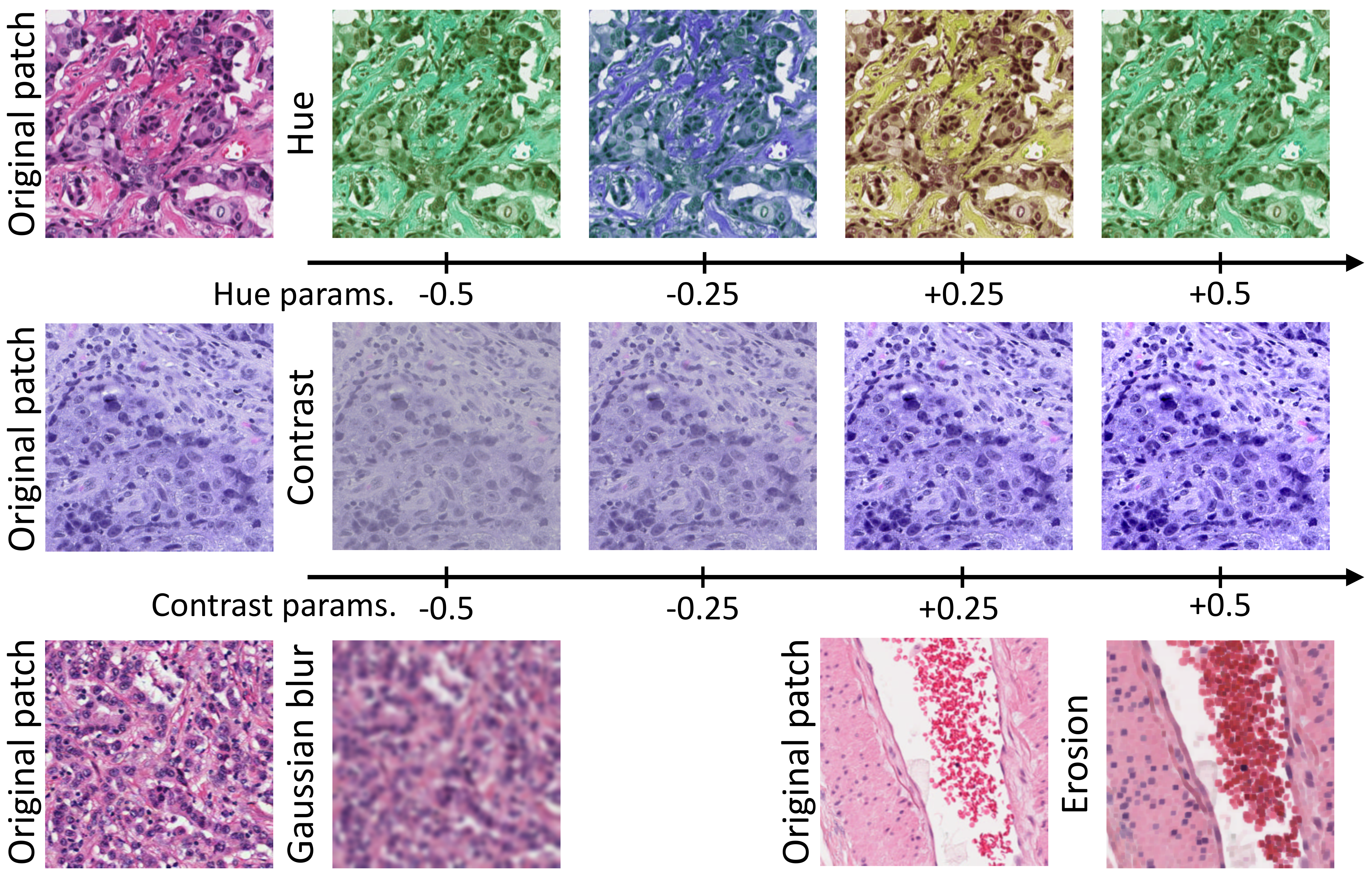}
    \caption{\textbf{Augmented image retrieval}: Visualization of the top-1 retrieved augmented images based on cosine similarity with \textit{HistAug} augmented features. As all retrieved images are correct, they correspond to the expected transformed patches for the given augmentation and parameter.
    }
    \label{fig:image_retrieval}
\end{figure}

\myparagraph{Augmented image retrieval} is used to visualize the generated features from \textit{HistAug}. We augment the  features from a patch with \textit{HistAug} given a transformation and associated parameter, and perform original image transformations on the same image patch for all considered augmentation types and parameter values. We then compute the cosine similarity between the generated features and the ones extracted from all the original image augmentations. As shown in Figure~\ref{fig:image_retrieval}, for different patches and augmentations (hue, contrast, gaussian blue, erosion), top-1 retrieved images from generated features are correct, showing that \textit{HistAug} properly simulates standard augmentations, even proper variations based on provided parameters.

\section{Conclusion}

Latent augmentation is promising in compute-demanding MIL training on small-scale datasets. We introduce \textit{HistAug}, a lightweight generator performing controllable augmentations in latent space. We improve MIL performance across models and datasets while being faster and less memory intensive than state-of-the-art diffusion-based counterparts. Additional studies demonstrate the ability of \textit{HistAug} to generalize to magnifications unseen during training, and its superiority against noise-based techniques. Qualitative studies highlight the controllability of our latent augmentations. Future work will study how our method can be extended to more types of augmentations, and how its controllability can further be improved.

\newpage

\myparagraph{Acknowledgments}  
This research was supported by the \textit{French National Research Agency} (\textit{ANR}) under project \textit{ANR-23-CE45-0029}, and the \textit{Health Data Hub} (\textit{HDH}) as part of the second edition of the \textit{France-Québec} call for projects \textit{Intelligence Artificielle en santé}. It was carried out using HPC resources from \textit{GENCI–IDRIS} (Grant \textit{2024-AD011015593}).

{
    \small
    \bibliographystyle{ieeenat_fullname}
    \bibliography{main}
}

\appendix
\setcounter{figure}{0}
\setcounter{table}{0}
\setcounter{section}{0}
\renewcommand{\thefigure}{S\arabic{figure}}
\renewcommand{\thetable}{S\arabic{table}}

\clearpage
\setcounter{page}{1}
\maketitlesupplementary

\begin{figure}[t]
  \centering

  \includegraphics[width=0.5\textwidth]{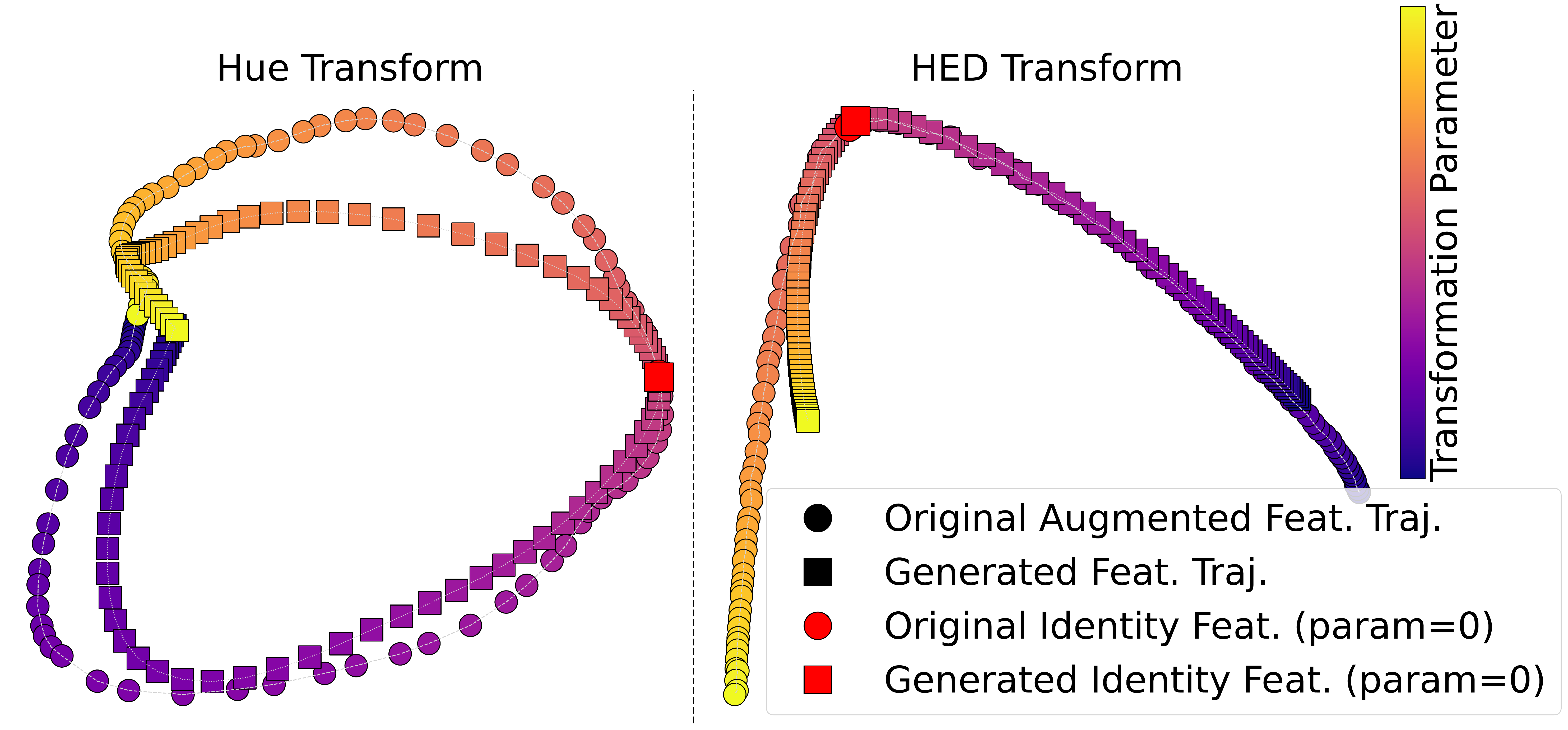}
  \caption{\textbf{Trajectories in augmentation space}: \textbf{Left} -- PCA trajectories of Hue transform for CONCH. \textbf{Right} -- PCA trajectories of HED transform for UNI. Circles represent the true augmented features extracted from the foundation model, while squares denote the features generated by our augmentation model.}
  \label{fig:pca_trajectories_supp_mat}
\end{figure}
\section{Additional details}

\subsection{Dataset preprocessing}
Slides are processed at $10\times$ (1 $\mu$m/pixel) and $20\times$ (0.5 $\mu$m/pixel) magnifications, with background regions removed and non-overlapping $256 \times 256$ pixel patches extracted from tissue regions using the CLAM toolbox \cite{lu2021data}. Patch features are then computed using the UNI \cite{chen2024towards} and CONCH \cite{lu2024visual} foundation models and stored for downstream analysis.

\subsection{Data splitting}

Since our generator $\rho$ is trained using TCGA data, and downstream tasks also rely on TCGA datasets, we carefully avoided data leakage as follows:
\begin{enumerate}
    \item We split the dataset into two subsets: a training portion (70\%), containing samples used to train the generator $\rho$, and a held-out portion (30\%).

    \item We generated five distinct training sets by bootstrapping (with replacement) from the 70\% training subset:
    \[
        \mathcal{D}_{\text{train}}^{(b)}, \quad b \in \{1,\dots,5\}.
    \]

    \item The remaining 30\% of held-out samples, which were never seen during generator training, were randomly partitioned into validation and test subsets five times (shuffle-split), creating:
    \[
        \mathcal{D}_{\text{val}}^{(b)}, \quad \mathcal{D}_{\text{test}}^{(b)}, \quad b \in \{1,\dots,5\}.
    \]
\end{enumerate}

This procedure resulted in five distinct dataset splits, each consisting of a training set $\mathcal{D}_{\text{train}}^{(b)}$, a validation set $\mathcal{D}_{\text{val}}^{(b)}$, and a test set $\mathcal{D}_{\text{test}}^{(b)}$, ensuring no sample overlap between the generator training and the validation/test sets used in downstream tasks.

\subsection{MIL training}
For both classification and survival tasks, models were trained for up to 200 epochs using the AdamW optimizer \cite{loshchilov2019decoupledweightdecayregularization} (except for the TransMIL model, which uses Lookahead RAdam \cite{zhang2019lookaheadoptimizerksteps}) with a learning rate of $10^{-4}$, weight decay of $10^{-5}$, and gradient accumulation over 4 steps. Early stopping with a patience of 30 epochs was applied in both cases to prevent overfitting.

Classification was optimized using the cross-entropy loss, while survival relied on the negative log-likelihood (NLL) loss. The best classification model was selected based on validation balanced accuracy, and for survival, based on the validation concordance index (C-index), both after the early stopping criterion was met.

During MIL training, feature-level augmentation was applied with a probability of 75\% across all augmentation strategies. For \textit{HistAug} and \textit{AugDiff}, augmentations were applied on-the-fly during training with a 75\% probability. For \textit{Patch Augmentation} (PAug), with the same probability, precomputed augmented features were used; otherwise, the original features were retained.

\subsection{Transformations}
In Table~\ref{tab:transforms_details} we present details of the stochastic image transformations applied, along with their respective parameter ranges.
\begin{table*}[t]
\centering
\caption{\textbf{Stochastic Image Transformations and Parameter Ranges}}
\small 
{
\setlength{\tabcolsep}{6pt}
\begin{tabular}{@{}lll@{}}
\toprule
\textbf{Transformation} & \textbf{Description} & \textbf{Sampling Range} \\ 
Crop       & Crop randomly from 4 corners or center &
                     Crop side size : \(\min(H,W) / 2\)  \\[2pt]

Dilation   & Morphological dilation &  Fixed \(4\times4\) kernel
                     \\[2pt]

Erosion    & Morphological erosion &  Fixed \(4\times4\) kernel \\[2pt]

Blur       & Gaussian blur & Fixed \(15\times15\) kernel \\[2pt]

Brightness & Brightness jitter & 
                     \(b \sim U[0.5,1.5]\) \\[2pt]

Contrast & Contrast jitter & 
                     \(c \sim U[0.5,1.5]\) \\[2pt]

Saturation & Saturation jitter & 
                     \(s \sim U[0.5,1.5]\) \\[2pt]

Hue & Hue jitter & 
                     \(h \sim U[-0.5, 0.5]\) \\[2pt]

HED        & \makecell[l]{Histology colour perturbation in HED space~\cite{pmlr-v143-faryna21a}. It separates hematoxylin, eosin, \\ and DAB channels to enable fine-grained, stain-aware perturbations.} &
                     - \\[2pt]

Flip       & Horizontal \emph{or} vertical flip &
                     - \\[2pt]

Rotate     & Rigid rotation & Angle \(\in\{90^{\circ},180^{\circ},270^{\circ}\}\) \\[2pt]

Gamma      & Power-law intensity transform &
                     \(\gamma \sim U[0.5, 1.5]\) \\ \bottomrule
\end{tabular}
\label{tab:transforms_details}
}
\end{table*}

\subsection{Augmented image retrieval}
Figure~\ref{fig:image_retrieval} in the main paper presents the results of a study we conduct to get insights about the underlying transformations simulated by \textit{HistAug} in the latent space. To this end, we sample patches and extract their embeddings with a foundation model, either \textit{UNI} or \textit{CONCH}. For each patch, we then predict the associated augmented embedding with \textit{HistAug} for a given transform and associated hyperparameter (e.g. hue transform with hyperparameter $-0.5$). Finally, we apply all considered transforms with, for each, a set of hyperparameter values (e.g. $\{-0.5, -0.25, +0.25, +0.25\}$ for hue and contrast) , to the original patch (standard image-space augmentation), and extract associated embeddings with the same foundation model as used previously. This gives us a query embedding, i.e. feature vector predicted by HistAug for one specific transformation, and a pool of key embeddings (37 in total), i.e. embeddings of augmented patches with all known transformations. We then perform image retrieval to find the top-1 key embedding which is the closest (based on cosine similarity) to the query embedding. Figure~\ref{fig:image_retrieval} thus displays the original patches and for each of them, the top-1 retrieved key augmented patch when generating the query embedding for different transforms (hue, contrast, gaussian blur, erosion) and associated hyperparameters ($\{-0.5, -0.25, +0.25, +0.25\}$ for hue and contrast).

\section{Trajectories in augmentation space}
\label{sec:figures}
Figure~\ref{fig:pca_trajectories_supp_mat} presents additional latent trajectory visualizations.

\begin{figure}[h]
  \centering
  \includegraphics[width=0.5\textwidth]{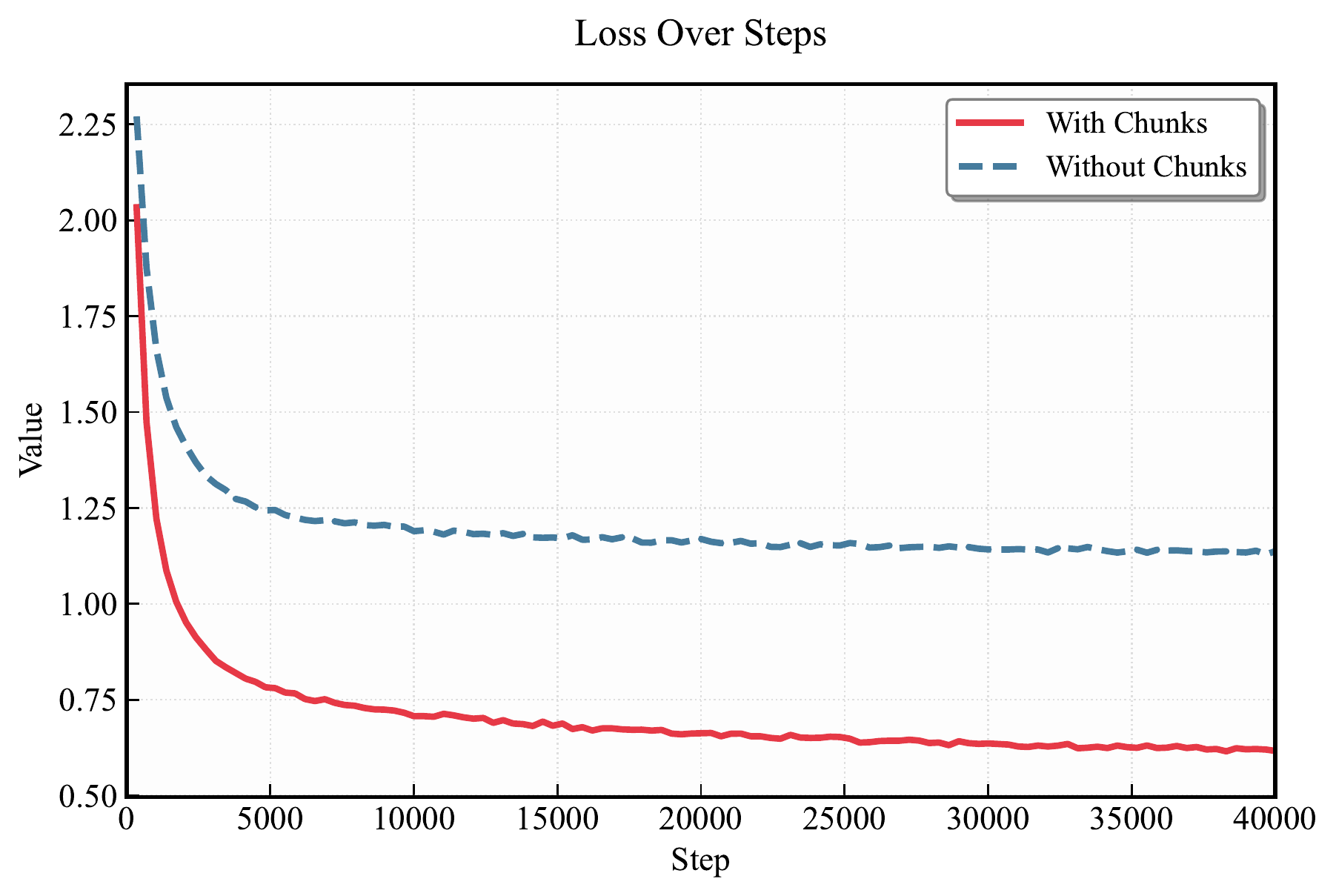}
  \caption{\textbf{Impact of chunking input features}: Chunking input features achieves lower training loss, reducing reconstruction error and leading to a faster training.}
  \label{fig:chunking_benefits}
\end{figure}
\section{Impact of chunking input features}
\label{sec:chunking}

Training a $d$-dimensional (e.g., $1024$) transformer directly on $\mathbf{z} \in \mathbb{R}^{1024}$ without chunking results in slower convergence and larger reconstruction errors. Indeed, as shown in Figure~\ref{fig:chunking_benefits},  chunking $\mathbf{z}$ into smaller segments and letting the transformer learn cross-chunk interactions achieves lower training MSE loss. We report the chunking details, including the original embedding dimension, the number of segments, and the resulting input dimension to the transformer network in Table~\ref{tab:chunking_for_extractors}.

\begin{table}[H]
\centering
\caption{\textbf{Chunking for UNI and CONCH Feature Extractors}}
\resizebox{\columnwidth}{!}{
\begin{tabular}{@{}c c c c@{}}
\toprule
\multirow{2}{*}{\textbf{Feature Extractor}} & \textbf{Orig. Embed.} & \textbf{Num. of} & \textbf{Transformer} \\
                                          & \textbf{Dimension}          & \textbf{Segments}  & \textbf{Input Dim}  \\ 
\midrule
UNI                                       & 1024                         & 8                           & 128 \\[2pt]
CONCH                                     & 512                          & 4                           & 128 \\[2pt]
\bottomrule
\end{tabular}
}
\label{tab:chunking_for_extractors}
\end{table}

\section{Evaluation on the BRACS Dataset for Breast Carcinoma Subtyping}

To assess the generalizability of our method beyond TCGA datasets, we report results on the BRACS dataset~\cite{bracs_dataset}, which focuses on multiclass classification for breast carcinoma subtyping. Specifically, we trained five MIL models (Abmil \cite{ilse2018attention}, CLAM variants \cite{lu2021data}, DSMIL \cite{li2021dual}, and TransMIL \cite{shao2021transmil})  on five different seeds each. For each MIL model, we computed the mean performance on the official test set by averaging the results across the five runs (with different seeds). The final reported values in Table~\ref{tab:bracs_res} represent the mean and standard deviation across these five MIL mean performances. Results show improvements of up to 3\% in both F1-score and balanced accuracy (Bacc), highlighting the effectiveness and generalizability of our approach beyond TCGA and on a multiclass setting.

\begin{table}[h]
\centering
\caption{\textbf{Additional results on BRACS dataset using 100\% of the training data}. The results are based on the official test set, aggregated across mean performance of five MIL models.}
\label{tab:bracs_res}
\begin{tabular}{llccc}
\toprule
& & \textbf{AUC} & \textbf{Bacc} & \textbf{F1 score} \\
\midrule
\multirow{3}{*}{\rotatebox[origin=c]{90}{\textbf{UNI}}}
  & Base        & 79.1$_{\pm2.6}$ & 38.5$_{\pm4.8}$ & 36.0$_{\pm4.7}$ \\
  & Ours(WSI)  & 79.7$_{\pm2.4}$ & 39.7$_{\pm5.0}$ & 37.5$_{\pm4.9}$ \\
  & Ours(Inst) & \textbf{80.3$_{\pm2.5}$} & \textbf{40.0$_{\pm5.2}$} & \textbf{37.6$_{\pm5.3}$} \\
\midrule
\multirow{3}{*}{\rotatebox[origin=c]{90}{\textbf{CONCH}}}
  & Base        & 82.3$_{\pm2.3}$ & 44.2$_{\pm3.5}$ & 41.9$_{\pm4.4}$ \\
  & Ours(WSI)  & \textbf{83.5$_{\pm2.3}$} & \textbf{47.4$_{\pm2.8}$} & \textbf{44.6$_{\pm3.2}$} \\
  & Ours(Inst) & 82.3$_{\pm2.2}$ & 44.8$_{\pm3.0}$ & 42.9$_{\pm3.6}$ \\
\bottomrule
\end{tabular}
\end{table}

\section{AugDiff Performance Scaling}
Despite its computational inefficiency, we trained AugDiff using the \textit{100\% training} setup on the BLCA dataset with the CONCH extractor—this being the only setting feasible to run within a reasonable timeframe. Notably, this experiment required approximately 500 GPU-hours for AugDiff, compared to only 5 GPU-hours for HistAug.  

As shown in Table~\ref{tab:augdiff_scaling}, HistAug continues to outperform AugDiff even when using the full training set. These results are consistent with the observations under the \textit{10\% training} setting reported in Table~\ref{tab:performance_accross_models} of the main paper, further validating the robustness and efficiency of our augmentation strategy.
\begin{table}[h]
\centering
\caption{Performance comparison on the BLCA dataset (CONCH extractor) using \textbf{100\% training data}. Mean ($\pm$ standard deviation) across 5 folds is reported. AugDiff is significantly more computationally expensive but still underperforms compared to our HistAug variants.}
\label{tab:augdiff_scaling}
\renewcommand{\arraystretch}{1}
\begin{adjustbox}{max width=0.75\linewidth}
\begin{tabular}{l|cccc}
\toprule
\textbf{Method} & Base & AugDiff & Ours (Inst) & Ours (WSI) \\
\midrule
\multirow{2}{*}{Abmil}   & 55.4 & \textbf{63.9} & 60.4 & 62.9 \\
                         & (\(\pm 3.0\)) & (\(\pm 4.0\)) & (\(\pm 5.0\)) & (\(\pm 3.0\)) \\[2pt]
\multirow{2}{*}{ClamMb}  & 58.7 & 63.4          & \textbf{64.4} & 62.6 \\
                         & (\(\pm 4.0\)) & (\(\pm 4.0\)) & (\(\pm 3.0\)) & (\(\pm 4.0\)) \\[2pt]
\multirow{2}{*}{ClamSb}  & 59.8 & 64.2          & \textbf{64.9} & 62.6 \\
                         & (\(\pm 4.0\)) & (\(\pm 3.0\)) & (\(\pm 4.0\)) & (\(\pm 2.0\)) \\[2pt]
\multirow{2}{*}{DSMIL}   & 55.6 & 61.4          & \textbf{64.0} & 61.9 \\
                         & (\(\pm 5.0\)) & (\(\pm 4.0\)) & (\(\pm 4.0\)) & (\(\pm 3.0\)) \\[2pt]
\multirow{2}{*}{TransMIL}& 60.4 & 60.0          & \textbf{63.6} & 62.7 \\
                         & (\(\pm 4.0\)) & (\(\pm 5.0\)) & (\(\pm 3.0\)) & (\(\pm 2.0\)) \\[2pt]
\midrule
\textbf{Mean}            & 58.0 & 62.6          & \textbf{63.5} & 62.5 \\
\bottomrule
\end{tabular}
\end{adjustbox}
\end{table}

\begin{table}[ht]
\centering
\caption{Mean cosine similarity (\%) between true augmented features and generated augmented features over 10 000 patches at 10X, alongside feature extractor invariance (\%). We also report the 95\% bootstrap confidence intervals (CI). The $\dagger$ symbole denotes out of training distribution.}
\label{tab:cosine_similarity_10X}
\begin{adjustbox}{width=0.5\textwidth}
\begin{tabular}{ll|cc|cc}
\toprule
  &  & \multicolumn{2}{c|}{\textbf{Feature Reconstruction}} & \multicolumn{2}{c}{\textbf{Feature Extractor Invariance}} \\
\cmidrule(lr){3-4} \cmidrule(lr){5-6}
  &  & \textbf{Cosine Sim (\%)} & \textbf{95\% CI} & \textbf{Cosine Sim (\%)} & \textbf{95\% CI} \\
\midrule

\multirow{6}{*}{\rotatebox[origin=c]{90}{\textbf{UNI}}}
& BLCA  & 81.0 & [80.9, 81.2] & 11.9 & [11.6, 12.1] \\
& BRCA  & 81.6 & [81.4, 81.7] & 11.9 & [11.7, 12.1] \\
& LUSC  & 81.1 & [80.9, 81.3] & 12.6 & [12.4, 12.8] \\
\cmidrule(lr){2-6}
& LUAD $\dagger$  & 80.3 & [80.2, 80.5] & 15.0 & [14.8, 15.3] \\
& UCEC $\dagger$  & 80.5 & [80.4, 80.7] &  9.5 & [9.3, 9.7] \\
& KIRC $\dagger$  & 80.5 & [80.4, 80.7] & 13.0 & [12.8, 13.2] \\
\midrule

\multirow{6}{*}{\rotatebox[origin=c]{90}{\textbf{CONCH}}}
& BLCA  & 90.3 & [90.2, 90.4] & 20.3 & [19.9, 20.7] \\
& BRCA  & 90.4 & [90.3, 90.5] & 24.3 & [24.0, 24.7] \\
& LUSC  & 90.4 & [90.3, 90.5] & 19.3 & [18.9, 19.6] \\
\cmidrule(lr){2-6}
& LUAD $\dagger$  & 90.2 & [90.1, 90.3] & 24.9 & [24.5, 25.3] \\
& UCEC $\dagger$  & 90.4 & [90.3, 90.5] & 19.7 & [19.3, 20.0] \\
& KIRC $\dagger$  & 89.9 & [89.8, 90.0] & 23.1 & [22.7, 23.5] \\

\bottomrule
\end{tabular}
\end{adjustbox}
\end{table}

\begin{table}[ht]
\centering
\caption{Mean Cosine similarity (\%) between true augmented features and generated augmented features over 10 000 patches at 20X, alongside feature extractor invariance (\%). We also report the 95\% bootstrap confidence intervals (CI). The $\dagger$ symbole denotes out of training distribution.}
\label{tab:cosine_similarity_20X}
\begin{adjustbox}{width=0.5\textwidth}
\begin{tabular}{ll|cc|cc}
\toprule
  &  & \multicolumn{2}{c|}{\textbf{Feature Reconstruction}} & \multicolumn{2}{c}{\textbf{Feature Extractor Invariance}} \\
\cmidrule(lr){3-4} \cmidrule(lr){5-6}
  &  & \textbf{Cosine Sim (\%)} & \textbf{95\% CI} & \textbf{Cosine Sim (\%)} & \textbf{95\% CI} \\
\midrule

\multirow{6}{*}{\rotatebox[origin=c]{90}{\textbf{UNI}}}
& BLCA & 74.9 & [74.7, 75.1] & 11.6 & [11.6, 11.8] \\
& BRCA  & 75.8 & [75.6, 76.0] & 11.7 & [11.5, 11.9] \\
& LUSC  & 74.9 & [74.7, 75.1] & 11.9 & [11.6, 12.0] \\
\cmidrule(lr){2-6}
& LUAD $\dagger$  & 74.6 & [74.4, 74.8] & 11.3 & [11.1, 11.5] \\
& UCEC $\dagger$  & 73.4 & [73.2, 73.5] & 12.9 & [12.7, 13.2] \\
& KIRC $\dagger$  & 72.1 & [71.9, 72.3] & 11.0 & [10.8, 11.1] \\
\midrule

\multirow{6}{*}{\rotatebox[origin=c]{90}{\textbf{CONCH}}}
& BLCA & 88.1 & [88.0, 88.3] & 27.7 & [27.4, 28.1] \\
& BRCA  & 88.6 & [88.5, 88.7] & 32.7 & [32.4, 33.1] \\
& LUSC  & 87.0 & [86.9, 87.1] & 27.7 & [27.4, 28.1] \\
\cmidrule(lr){2-6}
& LUAD $\dagger$  & 87.6 & [87.5, 87.7] & 23.7 & [23.3, 24.0] \\
& UCEC $\dagger$  & 88.5 & [88.4, 88.6] & 29.6 & [29.2, 29.9] \\
& KIRC $\dagger$  & 87.5 & [87.4, 87.6] & 29.8 & [29.5, 30.2] \\
\bottomrule
\end{tabular}
\end{adjustbox}
\end{table}

\section{Detailed result tables}
Tables~\ref{tab:cosine_similarity_10X} -~\ref{tab:100percent_20X} are detailed counterparts of tables presented in the main paper, showing more fine-grained results.

\subsection{Generator evaluation and and Feature Extractor Invariance}
Tables~\ref{tab:cosine_similarity_10X} and~\ref{tab:cosine_similarity_20X} present the mean cosine similarity between true and generated augmented features at 10X and 20X magnifications, alongside feature extractor invariance. The tables are the same as in the main paper, the main difference is that we include 95\% bootstrap confidence intervals.

\subsection{MIL evaluation at $10 \times$ magnification}
\myparagraph{Performance at 10\% Training Data} We compare several MIL architectures (Abmil \cite{ilse2018attention}, CLAM variants \cite{lu2021data}, DSMIL \cite{li2021dual}, and TransMIL \cite{shao2021transmil}) with different augmentation methods at 10\% training data. Results in Table~\ref{tab:10percent_10X} clearly indicate that instance-wise (Inst) and WSI-wise augmentation methods (WSI) consistently outperform the baseline, the diffusion-based augmentation (AugDiff), the noise-based augmentation (Noise) and the offline patch augmentation (PAug) across nearly all cancer types. Specifically, on the CONCH embeddings, our augmentation methods achieves substantial improvements in survival prediction (C-Index) and classification tasks (AUC) compared to baselines. For example, in BLCA, KIRC, and UCEC cancers, WSI-wise augmentation resulted in notable improvements of $3$–$6$ points in C-index compared to baseline models. The impact of noise perturbations at $10 \times$ magnification is also evaluated in these tables. Our learned augmentation strategies (Inst and WSI) consistently outperform the noise perturbation baseline across both 10\% and 100\% training data setups.

\myparagraph{Performance at 100\% Training Data} Using 100\% of the available training data (Table~\ref{tab:100percent_10X}), our augmentation methods enhance MIL performance in most cases over baselines across multiple cancer types. The improvements are more pronounced on the UNI embeddings, where instance-wise augmentation results in a mean improvement of approximately $4.5$ points in C-index for survival prediction tasks (e.g., BLCA improved from $54.5$ to $60.3$). For classification tasks (BRCA and NSCLC), WSI-wise augmentation methods maintain or slightly improve upon strong baseline performance.

\subsection{MIL evaluation at $20 \times$ magnification}

Tables~\ref{tab:10percent_20X} and~\ref{tab:100percent_20X} present the MIL evaluation results at 20X magnification, where we use our generator trained at 10X to generate augmented tiles. The trends observed at 10X remain consistent with WSI augmentations continuing to provide improvements. At 10\% training data (Table~\ref{tab:10percent_20X}), wsi-wise augmentation improves both survival prediction (C-Index) and classification (AUC), particularly for survival prediction. At 100\% training data (Table~\ref{tab:100percent_20X}), survival prediction still benefits from augmentation strategies. The results further demonstrate the generalizability of our method across magnifications.

\section{Code and Data Availability} 
The source code of our project will be made publicly available at \url{https://github.com/MICS-Lab/HistAug}. 

All TCGA datasets used in this study can be accessed via the Genomic Data Commons (GDC) portal at \url{https://portal.gdc.cancer.gov}. 

The BRACS dataset \cite{bracs_dataset} is publicly available at \url{https://www.bracs.icar.cnr.it}. 

The script for slide pre‑processing and patch extraction is available at \url{https://github.com/mahmoodlab/CLAM}. 

The code and pre‑trained weights for the UNI \cite{chen2024towards} and CONCH \cite{lu2024visual} models can be found on Hugging Face at \url{https://huggingface.co/MahmoodLab/UNI} and \url{https://huggingface.co/MahmoodLab/CONCH}, respectively.

\begin{table*}[ht]
\centering
\caption{\textbf{MIL evaluation with limited data}: Comparison at 10\% training data for 10X magnification. Survival (C-Index) on BLCA, KIRC, UCEC; classification (AUC) on BRCA, NSCLC. UNI (top), CONCH (bottom), \textbf{AugD}=AugDiff, \textbf{Noise}=feature-wise Gaussian noise, \textbf{PAug}=patch-wise augmentation, \textbf{Inst}=instance-wise augmentation (ours), \textbf{WSI}=wsi-wise augmentation (ours). Values are \%. Means ($\pm$ standard deviations) are reported over five splits.}
\label{tab:10percent_10X}
\begin{adjustbox}{width=1\textwidth}
\begin{tabular}{ll|cccccc|cccccc|cccccc|cccccc|cccccc}
\toprule
 &  & \multicolumn{18}{c|}{\textbf{Survival (C-Index)}} & \multicolumn{12}{c}{\textbf{Classification (AUC)}} \\
\cmidrule(lr){3-20}\cmidrule(lr){21-32}
 &  & \multicolumn{6}{c|}{BLCA} & \multicolumn{6}{c|}{KIRC} & \multicolumn{6}{c|}{UCEC} & \multicolumn{6}{c|}{BRCA} & \multicolumn{6}{c}{NSCLC} \\
\cmidrule(lr){3-8}\cmidrule(lr){9-14}\cmidrule(lr){15-20}\cmidrule(lr){21-26}\cmidrule(lr){27-32}
 & \textbf{Model} & \textbf{Base} & \textbf{AugD} & \textbf{Noise} & \textbf{PAug} & \textbf{Inst} & \textbf{WSI} & \textbf{Base} & \textbf{AugD} & \textbf{Noise} & \textbf{PAug} & \textbf{Inst} & \textbf{WSI} & \textbf{Base} & \textbf{AugD} & \textbf{Noise} & \textbf{PAug} & \textbf{Inst} & \textbf{WSI} & \textbf{Base} & \textbf{AugD} & \textbf{Noise} & \textbf{PAug} & \textbf{Inst} & \textbf{WSI} & \textbf{Base} & \textbf{AugD} & \textbf{Noise} & \textbf{PAug} & \textbf{Inst} & \textbf{WSI} \\
\midrule
\multirow{11}{*}{\rotatebox[origin=c]{90}{\textbf{UNI}}} & \multirow{2}{*}{Abmil} & 46.1 & 53.1 & 46.0 & 48.5 & 48.8 & 48.3 & 56.2 & 61.5 & 56.1 & 58.4 & 59.1 & 59.9 & 55.3 & 60.4 & 54.9 & 54.7 & 60.0 & 61.8 & 85.7 & 87.3 & 85.8 & 87.0 & 89.4 & 89.3 & 89.3 & 89.1 & 89.4 & 91.7 & 91.2 & 92.4 \\
 &  & (\(\pm 5.0\)) & (\(\pm 7.0\)) & (\(\pm 5.0\)) & (\(\pm 5.0\)) & (\(\pm 4.0\)) & (\(\pm 7.0\)) & (\(\pm 5.0\)) & (\(\pm 8.0\)) & (\(\pm 6.0\)) & (\(\pm 4.0\)) & (\(\pm 7.0\)) & (\(\pm 6.0\)) & (\(\pm 5.0\)) & (\(\pm 10.0\)) & (\(\pm 5.0\)) & (\(\pm 6.0\)) & (\(\pm 11.0\)) & (\(\pm 12.0\)) & (\(\pm 2.0\)) & (\(\pm 2.0\)) & (\(\pm 2.0\)) & (\(\pm 2.0\)) & (\(\pm 2.0\)) & (\(\pm 2.0\)) & (\(\pm 4.0\)) & (\(\pm 3.0\)) & (\(\pm 4.0\)) & (\(\pm 3.0\)) & (\(\pm 3.0\)) & (\(\pm 4.0\)) \\[2pt]
 & \multirow{2}{*}{ClamMb} & 49.2 & 48.4 & 49.4 & 47.9 & 50.8 & 49.6 & 59.0 & 67.2 & 58.9 & 57.3 & 64.5 & 64.3 & 57.5 & 62.6 & 57.5 & 59.5 & 62.1 & 60.6 & 87.0 & 87.2 & 87.0 & 89.5 & 89.4 & 88.9 & 90.9 & 87.4 & 91.7 & 92.7 & 92.6 & 93.7 \\
 &  & (\(\pm 6.0\)) & (\(\pm 7.0\)) & (\(\pm 6.0\)) & (\(\pm 5.0\)) & (\(\pm 7.0\)) & (\(\pm 8.0\)) & (\(\pm 9.0\)) & (\(\pm 3.0\)) & (\(\pm 9.0\)) & (\(\pm 6.0\)) & (\(\pm 5.0\)) & (\(\pm 5.0\)) & (\(\pm 6.0\)) & (\(\pm 8.0\)) & (\(\pm 6.0\)) & (\(\pm 9.0\)) & (\(\pm 7.0\)) & (\(\pm 12.0\)) & (\(\pm 2.0\)) & (\(\pm 2.0\)) & (\(\pm 2.0\)) & (\(\pm 2.0\)) & (\(\pm 2.0\)) & (\(\pm 2.0\)) & (\(\pm 4.0\)) & (\(\pm 5.0\)) & (\(\pm 3.0\)) & (\(\pm 4.0\)) & (\(\pm 3.0\)) & (\(\pm 2.0\)) \\[2pt]
 & \multirow{2}{*}{ClamSb} & 45.8 & 51.7 & 45.6 & 48.5 & 51.8 & 52.9 & 55.5 & 63.3 & 56.0 & 57.6 & 58.3 & 61.3 & 54.1 & 53.8 & 53.8 & 55.8 & 51.0 & 59.5 & 86.4 & 86.4 & 86.2 & 89.5 & 88.2 & 89.0 & 91.3 & 90.1 & 91.0 & 92.8 & 92.5 & 93.8 \\
 &  & (\(\pm 7.0\)) & (\(\pm 7.0\)) & (\(\pm 7.0\)) & (\(\pm 6.0\)) & (\(\pm 7.0\)) & (\(\pm 5.0\)) & (\(\pm 9.0\)) & (\(\pm 6.0\)) & (\(\pm 9.0\)) & (\(\pm 3.0\)) & (\(\pm 5.0\)) & (\(\pm 6.0\)) & (\(\pm 4.0\)) & (\(\pm 6.0\)) & (\(\pm 4.0\)) & (\(\pm 7.0\)) & (\(\pm 9.0\)) & (\(\pm 6.0\)) & (\(\pm 2.0\)) & (\(\pm 1.0\)) & (\(\pm 2.0\)) & (\(\pm 1.0\)) & (\(\pm 2.0\)) & (\(\pm 1.0\)) & (\(\pm 4.0\)) & (\(\pm 5.0\)) & (\(\pm 4.0\)) & (\(\pm 3.0\)) & (\(\pm 3.0\)) & (\(\pm 3.0\)) \\[2pt]
 & \multirow{2}{*}{DSMIL} & 46.2 & 46.7 & 46.3 & 46.3 & 50.3 & 50.8 & 63.3 & 67.5 & 61.4 & 63.4 & 64.5 & 66.5 & 63.0 & 67.2 & 62.7 & 66.8 & 64.2 & 64.8 & 85.2 & 73.2 & 87.3 & 87.0 & 86.9 & 87.0 & 81.2 & 79.3 & 87.0 & 84.3 & 87.2 & 85.6 \\
 &  & (\(\pm 3.0\)) & (\(\pm 4.0\)) & (\(\pm 4.0\)) & (\(\pm 6.0\)) & (\(\pm 5.0\)) & (\(\pm 3.0\)) & (\(\pm 8.0\)) & (\(\pm 3.0\)) & (\(\pm 8.0\)) & (\(\pm 8.0\)) & (\(\pm 4.0\)) & (\(\pm 4.0\)) & (\(\pm 2.0\)) & (\(\pm 4.0\)) & (\(\pm 4.0\)) & (\(\pm 5.0\)) & (\(\pm 7.0\)) & (\(\pm 6.0\)) & (\(\pm 4.0\)) & (\(\pm 2.0\)) & (\(\pm 2.0\)) & (\(\pm 3.0\)) & (\(\pm 3.0\)) & (\(\pm 3.0\)) & (\(\pm 3.0\)) & (\(\pm 4.0\)) & (\(\pm 4.0\)) & (\(\pm 3.0\)) & (\(\pm 3.0\)) & (\(\pm 2.0\)) \\[2pt]
 & \multirow{2}{*}{TransMIL} & 50.3 & 49.5 & 50.4 & 50.7 & 50.9 & 51.3 & 58.7 & 54.5 & 58.2 & 63.7 & 59.9 & 60.3 & 66.5 & 65.6 & 66.1 & 67.8 & 69.0 & 69.5 & 86.4 & 86.6 & 86.6 & 87.8 & 87.0 & 87.5 & 85.3 & 88.0 & 86.0 & 83.1 & 84.9 & 86.7 \\
 &  & (\(\pm 2.0\)) & (\(\pm 1.0\)) & (\(\pm 2.0\)) & (\(\pm 3.0\)) & (\(\pm 2.0\)) & (\(\pm 2.0\)) & (\(\pm 10.0\)) & (\(\pm 12.0\)) & (\(\pm 10.0\)) & (\(\pm 5.0\)) & (\(\pm 12.0\)) & (\(\pm 11.0\)) & (\(\pm 9.0\)) & (\(\pm 3.0\)) & (\(\pm 4.0\)) & (\(\pm 6.0\)) & (\(\pm 6.0\)) & (\(\pm 5.0\)) & (\(\pm 2.0\)) & (\(\pm 2.0\)) & (\(\pm 2.0\)) & (\(\pm 2.0\)) & (\(\pm 3.0\)) & (\(\pm 2.0\)) & (\(\pm 5.0\)) & (\(\pm 4.0\)) & (\(\pm 5.0\)) & (\(\pm 5.0\)) & (\(\pm 5.0\)) & (\(\pm 4.0\)) \\[2pt]
 & Mean & 47.5 & 49.9 & 47.5 & 48.4 & 50.5 & \textbf{50.6} & 58.5 & \textbf{62.8} & 58.1 & 60.1 & 61.3 & 62.5 & 59.3 & 61.9 & 59.0 & 60.9 & 61.3 & \textbf{63.2} & 86.1 & 84.1 & 86.6 & 88.2 & 88.2 & \textbf{88.3} & 87.6 & 86.8 & 89.0 & 88.9 & 89.7 & \textbf{90.4} \\
\midrule
\multirow{11}{*}{\rotatebox[origin=c]{90}{\textbf{CONCH}}} & \multirow{2}{*}{Abmil} & 49.7 & 52.2 & 51.6 & 51.0 & 52.3 & 52.2 & 60.4 & 65.8 & 61.3 & 61.6 & 65.3 & 67.6 & 56.1 & 61.3 & 55.6 & 64.2 & 63.2 & 65.8 & 91.6 & 91.9 & 90.2 & 88.2 & 90.9 & 91.7 & 95.2 & 93.8 & 93.0 & 94.2 & 94.5 & 95.0 \\
 &  & (\(\pm 5.0\)) & (\(\pm 5.0\)) & (\(\pm 3.0\)) & (\(\pm 4.0\)) & (\(\pm 7.0\)) & (\(\pm 8.0\)) & (\(\pm 5.0\)) & (\(\pm 3.0\)) & (\(\pm 6.0\)) & (\(\pm 11.0\)) & (\(\pm 9.0\)) & (\(\pm 8.0\)) & (\(\pm 5.0\)) & (\(\pm 6.0\)) & (\(\pm 6.0\)) & (\(\pm 6.0\)) & (\(\pm 9.0\)) & (\(\pm 2.0\)) & (\(\pm 2.0\)) & (\(\pm 1.0\)) & (\(\pm 2.0\)) & (\(\pm 4.0\)) & (\(\pm 1.0\)) & (\(\pm 2.0\)) & (\(\pm 1.0\)) & (\(\pm 2.0\)) & (\(\pm 2.0\)) & (\(\pm 1.0\)) & (\(\pm 2.0\)) & (\(\pm 2.0\)) \\[2pt]
 & \multirow{2}{*}{ClamMb} & 53.1 & 52.3 & 53.2 & 55.1 & 56.7 & 57.6 & 63.0 & 67.3 & 62.9 & 67.7 & 70.2 & 70.8 & 54.7 & 60.6 & 54.8 & 64.7 & 64.1 & 65.0 & 90.3 & 92.5 & 90.3 & 90.8 & 91.2 & 91.4 & 93.0 & 93.8 & 92.8 & 93.3 & 94.9 & 95.7 \\
 &  & (\(\pm 6.0\)) & (\(\pm 7.0\)) & (\(\pm 6.0\)) & (\(\pm 3.0\)) & (\(\pm 5.0\)) & (\(\pm 4.0\)) & (\(\pm 7.0\)) & (\(\pm 3.0\)) & (\(\pm 7.0\)) & (\(\pm 2.0\)) & (\(\pm 1.0\)) & (\(\pm 3.0\)) & (\(\pm 11.0\)) & (\(\pm 7.0\)) & (\(\pm 10.0\)) & (\(\pm 6.0\)) & (\(\pm 8.0\)) & (\(\pm 4.0\)) & (\(\pm 1.0\)) & (\(\pm 2.0\)) & (\(\pm 1.0\)) & (\(\pm 3.0\)) & (\(\pm 1.0\)) & (\(\pm 1.0\)) & (\(\pm 1.0\)) & (\(\pm 2.0\)) & (\(\pm 1.0\)) & (\(\pm 2.0\)) & (\(\pm 1.0\)) & (\(\pm 1.0\)) \\[2pt]
 & \multirow{2}{*}{ClamSb} & 48.2 & 50.7 & 48.6 & 53.6 & 53.8 & 49.7 & 65.6 & 66.0 & 66.2 & 65.4 & 70.5 & 72.0 & 58.5 & 62.3 & 58.6 & 65.3 & 64.4 & 63.3 & 90.2 & 92.3 & 90.6 & 88.1 & 90.6 & 91.4 & 92.5 & 94.6 & 92.4 & 94.1 & 94.5 & 95.6 \\
 &  & (\(\pm 4.0\)) & (\(\pm 7.0\)) & (\(\pm 4.0\)) & (\(\pm 5.0\)) & (\(\pm 5.0\)) & (\(\pm 6.0\)) & (\(\pm 5.0\)) & (\(\pm 3.0\)) & (\(\pm 5.0\)) & (\(\pm 9.0\)) & (\(\pm 4.0\)) & (\(\pm 6.0\)) & (\(\pm 10.0\)) & (\(\pm 7.0\)) & (\(\pm 10.0\)) & (\(\pm 5.0\)) & (\(\pm 11.0\)) & (\(\pm 5.0\)) & (\(\pm 1.0\)) & (\(\pm 1.0\)) & (\(\pm 1.0\)) & (\(\pm 3.0\)) & (\(\pm 1.0\)) & (\(\pm 2.0\)) & (\(\pm 3.0\)) & (\(\pm 2.0\)) & (\(\pm 2.0\)) & (\(\pm 2.0\)) & (\(\pm 2.0\)) & (\(\pm 1.0\)) \\[2pt]
 & \multirow{2}{*}{DSMIL} & 49.5 & 55.0 & 51.7 & 57.2 & 55.9 & 57.0 & 67.2 & 69.7 & 65.9 & 71.5 & 72.0 & 72.7 & 59.6 & 63.1 & 61.7 & 64.1 & 62.0 & 64.8 & 82.3 & 82.5 & 89.7 & 82.5 & 88.1 & 87.4 & 91.3 & 93.7 & 93.1 & 92.6 & 93.0 & 93.8 \\
 &  & (\(\pm 5.0\)) & (\(\pm 5.0\)) & (\(\pm 5.0\)) & (\(\pm 7.0\)) & (\(\pm 7.0\)) & (\(\pm 7.0\)) & (\(\pm 5.0\)) & (\(\pm 1.0\)) & (\(\pm 6.0\)) & (\(\pm 3.0\)) & (\(\pm 3.0\)) & (\(\pm 3.0\)) & (\(\pm 9.0\)) & (\(\pm 9.0\)) & (\(\pm 8.0\)) & (\(\pm 6.0\)) & (\(\pm 8.0\)) & (\(\pm 5.0\)) & (\(\pm 3.0\)) & (\(\pm 4.0\)) & (\(\pm 2.0\)) & (\(\pm 3.0\)) & (\(\pm 3.0\)) & (\(\pm 2.0\)) & (\(\pm 2.0\)) & (\(\pm 2.0\)) & (\(\pm 2.0\)) & (\(\pm 1.0\)) & (\(\pm 2.0\)) & (\(\pm 2.0\)) \\[2pt]
 & \multirow{2}{*}{TransMIL} & 53.7 & 55.0 & 53.5 & 53.9 & 53.9 & 53.8 & 59.3 & 60.8 & 59.5 & 60.5 & 64.1 & 64.7 & 64.0 & 62.0 & 63.3 & 64.2 & 64.6 & 65.5 & 91.4 & 91.2 & 91.4 & 90.9 & 91.2 & 92.1 & 92.0 & 93.0 & 92.3 & 92.0 & 92.7 & 93.0 \\
 &  & (\(\pm 4.0\)) & (\(\pm 6.0\)) & (\(\pm 4.0\)) & (\(\pm 4.0\)) & (\(\pm 5.0\)) & (\(\pm 5.0\)) & (\(\pm 9.0\)) & (\(\pm 8.0\)) & (\(\pm 10.0\)) & (\(\pm 8.0\)) & (\(\pm 6.0\)) & (\(\pm 5.0\)) & (\(\pm 7.0\)) & (\(\pm 9.0\)) & (\(\pm 8.0\)) & (\(\pm 9.0\)) & (\(\pm 7.0\)) & (\(\pm 5.0\)) & (\(\pm 1.0\)) & (\(\pm 1.0\)) & (\(\pm 1.0\)) & (\(\pm 2.0\)) & (\(\pm 2.0\)) & (\(\pm 1.0\)) & (\(\pm 2.0\)) & (\(\pm 2.0\)) & (\(\pm 2.0\)) & (\(\pm 2.0\)) & (\(\pm 1.0\)) & (\(\pm 2.0\)) \\[2pt]
 & Mean & 50.8 & 53.0 & 51.7 & 54.2 & \textbf{54.5} & 54.1 & 63.1 & 65.9 & 63.2 & 65.3 & 68.4 & \textbf{69.6} & 58.6 & 61.9 & 58.8 & 64.5 & 63.7 & \textbf{64.9} & 89.2 & 90.1 & 90.4 & 88.1 & 90.4 & \textbf{90.8} & 92.8 & 93.8 & 92.7 & 93.2 & 93.9 & \textbf{94.6} \\
\bottomrule
\end{tabular}
\end{adjustbox}
\end{table*}

\begin{table*}[ht]
\centering
\caption{\textbf{MIL evaluation with full data}: Results at 100\% training data for 10X magnification. Survival (C-Index) on BLCA, KIRC, UCEC; classification (AUC) on BRCA, NSCLC. UNI (top), CONCH (bottom), \textbf{Noise}=feature-wise Gaussian noise, \textbf{PAug}=patch-wise augmentation, \textbf{Inst}=instance-wise augmentation (ours), \textbf{WSI}=wsi-wise augmentation (ours). Values are \%. Means ($\pm$ standard deviations) are reported over five splits.}
\label{tab:100percent_10X}
\begin{adjustbox}{width=1\textwidth}
\begin{tabular}{ll|ccccc|ccccc|ccccc|ccccc|ccccc}
\toprule
 &  & \multicolumn{15}{c|}{\textbf{Survival (C-Index)}} & \multicolumn{10}{c}{\textbf{Classification (AUC)}} \\
\cmidrule(lr){3-17}\cmidrule(lr){18-27}
 &  & \multicolumn{5}{c|}{BLCA} & \multicolumn{5}{c|}{KIRC} & \multicolumn{5}{c|}{UCEC} & \multicolumn{5}{c|}{BRCA} & \multicolumn{5}{c}{NSCLC} \\
\cmidrule(lr){3-7}\cmidrule(lr){8-12}\cmidrule(lr){13-17}\cmidrule(lr){18-22}\cmidrule(lr){23-27}
 & \textbf{Model} & \textbf{Base} & \textbf{Noise} & \textbf{PAug} & \textbf{Inst} & \textbf{WSI} & \textbf{Base} & \textbf{Noise} & \textbf{PAug} & \textbf{Inst} & \textbf{WSI} & \textbf{Base} & \textbf{Noise} & \textbf{PAug} & \textbf{Inst} & \textbf{WSI} & \textbf{Base} & \textbf{Noise} & \textbf{PAug} & \textbf{Inst} & \textbf{WSI} & \textbf{Base} & \textbf{Noise} & \textbf{PAug} & \textbf{Inst} & \textbf{WSI} \\
\midrule
\multirow{11}{*}{\rotatebox[origin=c]{90}{\textbf{UNI}}} & \multirow{2}{*}{Abmil} & 51.8 & 54.1 & 59.2 & 59.1 & 57.4 & 67.1 & 66.5 & 68.5 & 68.6 & 69.1 & 63.0 & 63.5 & 63.0 & 61.5 & 62.7 & 93.0 & 93.3 & 93.8 & 92.1 & 92.1 & 97.8 & 98.0 & 93.5 & 97.9 & 97.5 \\
 &  & (\(\pm 6.0\)) & (\(\pm 5.0\)) & (\(\pm 5.0\)) & (\(\pm 4.0\)) & (\(\pm 6.0\)) & (\(\pm 3.0\)) & (\(\pm 5.0\)) & (\(\pm 3.0\)) & (\(\pm 5.0\)) & (\(\pm 4.0\)) & (\(\pm 9.0\)) & (\(\pm 10.0\)) & (\(\pm 3.0\)) & (\(\pm 12.0\)) & (\(\pm 5.0\)) & (\(\pm 2.0\)) & (\(\pm 2.0\)) & (\(\pm 2.0\)) & (\(\pm 2.0\)) & (\(\pm 2.0\)) & (\(\pm 1.0\)) & (\(\pm 1.0\)) & (\(\pm 2.0\)) & (\(\pm 1.0\)) & (\(\pm 1.0\)) \\[2pt]
 & \multirow{2}{*}{ClamMb} & 58.0 & 56.7 & 55.2 & 56.7 & 56.7 & 64.0 & 64.9 & 68.4 & 67.3 & 69.1 & 65.5 & 65.1 & 62.2 & 61.6 & 62.2 & 93.0 & 93.1 & 93.0 & 92.5 & 92.4 & 98.6 & 98.1 & 94.1 & 97.1 & 97.5 \\
 &  & (\(\pm 7.0\)) & (\(\pm 8.0\)) & (\(\pm 4.0\)) & (\(\pm 6.0\)) & (\(\pm 4.0\)) & (\(\pm 3.0\)) & (\(\pm 4.0\)) & (\(\pm 2.0\)) & (\(\pm 3.0\)) & (\(\pm 3.0\)) & (\(\pm 5.0\)) & (\(\pm 3.0\)) & (\(\pm 4.0\)) & (\(\pm 11.0\)) & (\(\pm 5.0\)) & (\(\pm 2.0\)) & (\(\pm 2.0\)) & (\(\pm 1.0\)) & (\(\pm 2.0\)) & (\(\pm 2.0\)) & (\(\pm 1.0\)) & (\(\pm 1.0\)) & (\(\pm 1.0\)) & (\(\pm 1.0\)) & (\(\pm 2.0\)) \\[2pt]
 & \multirow{2}{*}{ClamSb} & 55.2 & 52.5 & 58.7 & 62.8 & 63.9 & 69.9 & 69.0 & 65.3 & 68.5 & 67.8 & 56.2 & 57.4 & 65.3 & 69.4 & 63.3 & 93.5 & 92.8 & 93.2 & 92.3 & 92.5 & 98.5 & 98.3 & 94.4 & 97.6 & 97.4 \\
 &  & (\(\pm 5.0\)) & (\(\pm 5.0\)) & (\(\pm 3.0\)) & (\(\pm 2.0\)) & (\(\pm 2.0\)) & (\(\pm 5.0\)) & (\(\pm 7.0\)) & (\(\pm 3.0\)) & (\(\pm 5.0\)) & (\(\pm 4.0\)) & (\(\pm 7.0\)) & (\(\pm 3.0\)) & (\(\pm 10.0\)) & (\(\pm 6.0\)) & (\(\pm 11.0\)) & (\(\pm 2.0\)) & (\(\pm 2.0\)) & (\(\pm 2.0\)) & (\(\pm 2.0\)) & (\(\pm 3.0\)) & (\(\pm 1.0\)) & (\(\pm 1.0\)) & (\(\pm 2.0\)) & (\(\pm 1.0\)) & (\(\pm 1.0\)) \\[2pt]
 & \multirow{2}{*}{DSMIL} & 48.8 & 58.2 & 50.1 & 60.3 & 58.5 & 64.4 & 66.1 & 64.9 & 69.0 & 68.8 & 65.3 & 60.1 & 64.6 & 69.2 & 68.9 & 90.9 & 92.5 & 93.4 & 92.7 & 92.3 & 98.1 & 97.4 & 91.9 & 98.0 & 98.4 \\
 &  & (\(\pm 5.0\)) & (\(\pm 4.0\)) & (\(\pm 6.0\)) & (\(\pm 3.0\)) & (\(\pm 4.0\)) & (\(\pm 7.0\)) & (\(\pm 4.0\)) & (\(\pm 3.0\)) & (\(\pm 4.0\)) & (\(\pm 4.0\)) & (\(\pm 7.0\)) & (\(\pm 14.0\)) & (\(\pm 3.0\)) & (\(\pm 6.0\)) & (\(\pm 6.0\)) & (\(\pm 2.0\)) & (\(\pm 2.0\)) & (\(\pm 2.0\)) & (\(\pm 1.0\)) & (\(\pm 2.0\)) & (\(\pm 0.0\)) & (\(\pm 1.0\)) & (\(\pm 1.0\)) & (\(\pm 1.0\)) & (\(\pm 1.0\)) \\[2pt]
 & \multirow{2}{*}{TransMIL} & 58.7 & 58.3 & 60.4 & 62.5 & 60.9 & 63.5 & 63.0 & 67.4 & 64.8 & 66.3 & 68.9 & 68.3 & 68.6 & 65.6 & 66.9 & 92.9 & 92.8 & 93.3 & 92.4 & 92.9 & 95.6 & 95.9 & 93.6 & 95.8 & 97.8 \\
 &  & (\(\pm 3.0\)) & (\(\pm 2.0\)) & (\(\pm 4.0\)) & (\(\pm 3.0\)) & (\(\pm 2.0\)) & (\(\pm 3.0\)) & (\(\pm 3.0\)) & (\(\pm 3.0\)) & (\(\pm 3.0\)) & (\(\pm 3.0\)) & (\(\pm 7.0\)) & (\(\pm 9.0\)) & (\(\pm 4.0\)) & (\(\pm 5.0\)) & (\(\pm 9.0\)) & (\(\pm 2.0\)) & (\(\pm 2.0\)) & (\(\pm 2.0\)) & (\(\pm 2.0\)) & (\(\pm 3.0\)) & (\(\pm 1.0\)) & (\(\pm 1.0\)) & (\(\pm 2.0\)) & (\(\pm 1.0\)) & (\(\pm 0.0\)) \\[2pt]
 & Mean & 54.5 & 56.0 & 56.7 & \textbf{60.3} & 59.5 & 65.8 & 65.9 & 66.9 & 67.6 & \textbf{68.2} & 63.8 & 62.9 & 64.7 & \textbf{65.5} & 64.8 & 92.7 & 92.9 & \textbf{93.3} & 92.4 & 92.4 & \textbf{97.7} & 97.5 & 93.5 & 97.3 & \textbf{97.7} \\
\midrule
\multirow{11}{*}{\rotatebox[origin=c]{90}{\textbf{CONCH}}} & \multirow{2}{*}{Abmil} & 55.4 & 56.6 & 60.9 & 60.4 & 62.9 & 67.6 & 72.8 & 69.6 & 69.9 & 71.6 & 59.3 & 59.6 & 66.9 & 65.3 & 67.3 & 93.8 & 93.0 & 92.7 & 93.3 & 94.0 & 98.0 & 97.3 & 94.8 & 98.4 & 98.4 \\
 &  & (\(\pm 3.0\)) & (\(\pm 3.0\)) & (\(\pm 3.0\)) & (\(\pm 5.0\)) & (\(\pm 3.0\)) & (\(\pm 4.0\)) & (\(\pm 2.0\)) & (\(\pm 3.0\)) & (\(\pm 1.0\)) & (\(\pm 2.0\)) & (\(\pm 4.0\)) & (\(\pm 6.0\)) & (\(\pm 9.0\)) & (\(\pm 12.0\)) & (\(\pm 12.0\)) & (\(\pm 1.0\)) & (\(\pm 2.0\)) & (\(\pm 2.0\)) & (\(\pm 1.0\)) & (\(\pm 2.0\)) & (\(\pm 2.0\)) & (\(\pm 1.0\)) & (\(\pm 1.0\)) & (\(\pm 1.0\)) & (\(\pm 1.0\)) \\[2pt]
 & \multirow{2}{*}{ClamMb} & 58.7 & 58.7 & 62.7 & 64.4 & 62.6 & 70.0 & 69.9 & 71.6 & 71.9 & 73.2 & 62.1 & 63.9 & 71.8 & 67.0 & 67.0 & 94.3 & 94.2 & 92.8 & 93.5 & 93.7 & 98.4 & 98.1 & 95.3 & 98.4 & 98.5 \\
 &  & (\(\pm 4.0\)) & (\(\pm 5.0\)) & (\(\pm 4.0\)) & (\(\pm 3.0\)) & (\(\pm 4.0\)) & (\(\pm 5.0\)) & (\(\pm 4.0\)) & (\(\pm 2.0\)) & (\(\pm 3.0\)) & (\(\pm 3.0\)) & (\(\pm 7.0\)) & (\(\pm 7.0\)) & (\(\pm 10.0\)) & (\(\pm 8.0\)) & (\(\pm 4.0\)) & (\(\pm 2.0\)) & (\(\pm 2.0\)) & (\(\pm 2.0\)) & (\(\pm 2.0\)) & (\(\pm 2.0\)) & (\(\pm 1.0\)) & (\(\pm 1.0\)) & (\(\pm 1.0\)) & (\(\pm 1.0\)) & (\(\pm 1.0\)) \\[2pt]
 & \multirow{2}{*}{ClamSb} & 59.8 & 58.6 & 63.1 & 64.9 & 62.6 & 66.6 & 66.5 & 71.2 & 70.2 & 70.9 & 60.6 & 59.0 & 65.0 & 62.9 & 63.0 & 94.1 & 92.8 & 92.6 & 93.9 & 94.2 & 97.5 & 95.9 & 95.0 & 97.7 & 98.0 \\
 &  & (\(\pm 4.0\)) & (\(\pm 3.0\)) & (\(\pm 4.0\)) & (\(\pm 4.0\)) & (\(\pm 2.0\)) & (\(\pm 7.0\)) & (\(\pm 4.0\)) & (\(\pm 5.0\)) & (\(\pm 1.0\)) & (\(\pm 2.0\)) & (\(\pm 13.0\)) & (\(\pm 11.0\)) & (\(\pm 10.0\)) & (\(\pm 9.0\)) & (\(\pm 3.0\)) & (\(\pm 2.0\)) & (\(\pm 2.0\)) & (\(\pm 2.0\)) & (\(\pm 2.0\)) & (\(\pm 2.0\)) & (\(\pm 2.0\)) & (\(\pm 1.0\)) & (\(\pm 1.0\)) & (\(\pm 1.0\)) & (\(\pm 1.0\)) \\[2pt]
 & \multirow{2}{*}{DSMIL} & 55.6 & 61.9 & 59.0 & 64.0 & 61.9 & 67.8 & 69.0 & 69.1 & 70.6 & 72.2 & 58.1 & 62.5 & 63.9 & 64.6 & 64.8 & 89.8 & 93.9 & 92.5 & 94.4 & 93.9 & 98.2 & 97.5 & 95.4 & 98.2 & 97.5 \\
 &  & (\(\pm 5.0\)) & (\(\pm 5.0\)) & (\(\pm 7.0\)) & (\(\pm 4.0\)) & (\(\pm 3.0\)) & (\(\pm 5.0\)) & (\(\pm 7.0\)) & (\(\pm 2.0\)) & (\(\pm 4.0\)) & (\(\pm 4.0\)) & (\(\pm 8.0\)) & (\(\pm 10.0\)) & (\(\pm 6.0\)) & (\(\pm 9.0\)) & (\(\pm 5.0\)) & (\(\pm 2.0\)) & (\(\pm 2.0\)) & (\(\pm 2.0\)) & (\(\pm 2.0\)) & (\(\pm 2.0\)) & (\(\pm 1.0\)) & (\(\pm 1.0\)) & (\(\pm 1.0\)) & (\(\pm 1.0\)) & (\(\pm 2.0\)) \\[2pt]
 & \multirow{2}{*}{TransMIL} & 60.4 & 60.2 & 60.6 & 63.6 & 62.7 & 70.7 & 70.1 & 68.0 & 68.1 & 68.1 & 60.2 & 60.0 & 56.4 & 58.6 & 64.1 & 93.0 & 93.0 & 91.3 & 92.2 & 92.7 & 97.6 & 96.2 & 92.5 & 97.8 & 96.8 \\
 &  & (\(\pm 4.0\)) & (\(\pm 4.0\)) & (\(\pm 5.0\)) & (\(\pm 3.0\)) & (\(\pm 2.0\)) & (\(\pm 4.0\)) & (\(\pm 4.0\)) & (\(\pm 5.0\)) & (\(\pm 2.0\)) & (\(\pm 4.0\)) & (\(\pm 9.0\)) & (\(\pm 9.0\)) & (\(\pm 4.0\)) & (\(\pm 13.0\)) & (\(\pm 10.0\)) & (\(\pm 2.0\)) & (\(\pm 2.0\)) & (\(\pm 1.0\)) & (\(\pm 2.0\)) & (\(\pm 2.0\)) & (\(\pm 1.0\)) & (\(\pm 1.0\)) & (\(\pm 1.0\)) & (\(\pm 1.0\)) & (\(\pm 1.0\)) \\[2pt]
 & Mean & 58.0 & 59.2 & 61.3 & \textbf{63.5} & 62.5 & 68.5 & 69.7 & 69.9 & 70.1 & \textbf{71.2} & 60.1 & 61.0 & 64.8 & 63.7 & \textbf{65.2} & 93.0 & 93.4 & 92.4 & 93.5 & \textbf{93.7} & 97.9 & 97.0 & 94.6 & \textbf{98.1} & 97.8 \\
\bottomrule
\end{tabular}
\end{adjustbox}
\end{table*}

\begin{table*}[ht]
\centering
\caption{Comparison at 10\% training data (20X magnification). %
         \textbf{UNI} (top) and \textbf{CONCH} (bottom). %
         Values are reported in percentage. Means ($\pm$ standard deviations) are reported over five splits.}
\label{tab:10percent_20X}
\begin{adjustbox}{max width=\textwidth, center}
\scriptsize
\begin{tabular}{ll|cc|cc|cc|cc|cc}
\toprule
 &  & \multicolumn{6}{c|}{\textbf{Survival (C-Index)}} & \multicolumn{4}{c}{\textbf{Classification (AUC)}} \\
\cmidrule(lr){3-8}\cmidrule(lr){9-12}
 &  & \multicolumn{2}{c|}{BLCA} & \multicolumn{2}{c|}{KIRC} & \multicolumn{2}{c|}{UCEC} & \multicolumn{2}{c|}{BRCA} & \multicolumn{2}{c}{NSCLC} \\
\cmidrule(lr){3-4}\cmidrule(lr){5-6}\cmidrule(lr){7-8}\cmidrule(lr){9-10}\cmidrule(lr){11-12}
 & Model & Base & WSI & Base & WSI & Base & WSI & Base & WSI & Base & WSI \\
\midrule
\multirow{12}{*}{\rotatebox[origin=c]{90}{\textbf{UNI}}}& \multirow{2}{*}{Abmil} & 45.8 & 48.5 & 55.3 & 58.5 & 54.6 & 58.8 & 86.4 & 88.1 & 90.9 & 89.7 \\
&  & ($\pm 6.0$) & ($\pm 6.0$) & ($\pm 3.0$) & ($\pm 5.0$) & ($\pm 4.0$) & ($\pm 6.0$) & ($\pm 3.0$) & ($\pm 2.0$) & ($\pm 4.0$) & ($\pm 7.0$) \\[3pt]
& \multirow{2}{*}{ClamMb} & 45.7 & 49.1 & 56.4 & 61.4 & 51.6 & 55.5 & 87.1 & 88.5 & 91.9 & 92.8 \\
&  & ($\pm 7.0$) & ($\pm 9.0$) & ($\pm 6.0$) & ($\pm 6.0$) & ($\pm 6.0$) & ($\pm 8.0$) & ($\pm 2.0$) & ($\pm 1.0$) & ($\pm 4.0$) & ($\pm 4.0$) \\[3pt]
& \multirow{2}{*}{ClamSb} & 46.3 & 50.7 & 56.6 & 57.3 & 52.8 & 56.4 & 87.0 & 87.5 & 92.1 & 93.6 \\
&  & ($\pm 6.0$) & ($\pm 7.0$) & ($\pm 5.0$) & ($\pm 5.0$) & ($\pm 5.0$) & ($\pm 3.0$) & ($\pm 2.0$) & ($\pm 2.0$) & ($\pm 4.0$) & ($\pm 4.0$) \\[3pt]
& \multirow{2}{*}{DSMIL} & 45.6 & 50.2 & 58.3 & 63.6 & 56.9 & 60.0 & 81.6 & 85.2 & 81.3 & 84.1 \\
&  & ($\pm 7.0$) & ($\pm 8.0$) & ($\pm 7.0$) & ($\pm 2.0$) & ($\pm 4.0$) & ($\pm 3.0$) & ($\pm 3.0$) & ($\pm 2.0$) & ($\pm 6.0$) & ($\pm 4.0$) \\[3pt]
& \multirow{2}{*}{TransMIL} & 51.5 & 51.1 & 61.5 & 65.3 & 65.2 & 68.4 & 85.5 & 86.7 & 83.5 & 83.6 \\
&  & ($\pm 7.0$) & ($\pm 4.0$) & ($\pm 4.0$) & ($\pm 3.0$) & ($\pm 5.0$) & ($\pm 4.0$) & ($\pm 1.0$) & ($\pm 2.0$) & ($\pm 7.0$) & ($\pm 7.0$) \\[3pt]
& \textbf{Mean} & 47.0 & \textbf{49.9} & 57.6 & \textbf{61.2} & 56.2 & \textbf{59.8} & 85.5 & \textbf{87.2} & 87.9 & \textbf{88.8} \\
\midrule
\multirow{12}{*}{\rotatebox[origin=c]{90}{\textbf{CONCH}}}& \multirow{2}{*}{Abmil} & 50.4 & 52.1 & 59.0 & 65.7 & 55.9 & 65.4 & 90.1 & 90.7 & 92.8 & 95.0 \\
&  & ($\pm 6.0$) & ($\pm 6.0$) & ($\pm 7.0$) & ($\pm 9.0$) & ($\pm 8.0$) & ($\pm 10.0$) & ($\pm 4.0$) & ($\pm 3.0$) & ($\pm 4.0$) & ($\pm 3.0$) \\[3pt]
& \multirow{2}{*}{ClamMb} & 51.7 & 55.4 & 64.9 & 70.7 & 60.2 & 59.2 & 90.2 & 90.9 & 92.9 & 95.9 \\
&  & ($\pm 7.0$) & ($\pm 3.0$) & ($\pm 6.0$) & ($\pm 3.0$) & ($\pm 10.0$) & ($\pm 11.0$) & ($\pm 3.0$) & ($\pm 1.0$) & ($\pm 4.0$) & ($\pm 2.0$) \\[3pt]
& \multirow{2}{*}{ClamSb} & 46.5 & 53.9 & 66.4 & 70.2 & 57.9 & 62.5 & 90.9 & 90.5 & 92.3 & 95.4 \\
&  & ($\pm 7.0$) & ($\pm 6.0$) & ($\pm 5.0$) & ($\pm 6.0$) & ($\pm 9.0$) & ($\pm 11.0$) & ($\pm 3.0$) & ($\pm 2.0$) & ($\pm 5.0$) & ($\pm 3.0$) \\[3pt]
& \multirow{2}{*}{DSMIL} & 50.2 & 57.7 & 69.8 & 72.4 & 60.5 & 62.3 & 83.0 & 83.6 & 92.4 & 94.6 \\
&  & ($\pm 5.0$) & ($\pm 6.0$) & ($\pm 3.0$) & ($\pm 4.0$) & ($\pm 9.0$) & ($\pm 10.0$) & ($\pm 3.0$) & ($\pm 5.0$) & ($\pm 3.0$) & ($\pm 2.0$) \\[3pt]
& \multirow{2}{*}{TransMIL} & 53.1 & 55.4 & 62.1 & 64.8 & 66.0 & 66.2 & 89.5 & 90.4 & 92.2 & 94.5 \\
&  & ($\pm 7.0$) & ($\pm 8.0$) & ($\pm 8.0$) & ($\pm 6.0$) & ($\pm 8.0$) & ($\pm 8.0$) & ($\pm 3.0$) & ($\pm 1.0$) & ($\pm 4.0$) & ($\pm 2.0$) \\[3pt]
& \textbf{Mean} & 50.4 & \textbf{54.9} & 64.4 & \textbf{68.8} & 60.1 & \textbf{63.1} & 88.7 & \textbf{89.2} & 92.5 & \textbf{95.1} \\

\bottomrule
\end{tabular}
\end{adjustbox}
\end{table*}

\begin{table*}[ht]
\centering
\caption{Comparison at 100\% training data (20X magnification). %
         \textbf{UNI} (top) and \textbf{CONCH} (bottom). %
         Values are reported in percentage. Means ($\pm$ standard deviations) are reported over five splits.}
\label{tab:100percent_20X}
\begin{adjustbox}{max width=\textwidth, center}
\scriptsize
\begin{tabular}{ll|cc|cc|cc|cc|cc}
\toprule
 &  & \multicolumn{6}{c|}{\textbf{Survival (C-Index)}} & \multicolumn{4}{c}{\textbf{Classification (AUC)}} \\
\cmidrule(lr){3-8}\cmidrule(lr){9-12}
 &  & \multicolumn{2}{c|}{BLCA} & \multicolumn{2}{c|}{KIRC} & \multicolumn{2}{c|}{UCEC} & \multicolumn{2}{c|}{BRCA} & \multicolumn{2}{c}{NSCLC} \\
\cmidrule(lr){3-4}\cmidrule(lr){5-6}\cmidrule(lr){7-8}\cmidrule(lr){9-10}\cmidrule(lr){11-12}
 & Model & Base & WSI & Base & WSI & Base & WSI & Base & WSI & Base & WSI \\
\midrule
\multirow{12}{*}{\rotatebox[origin=c]{90}{\textbf{UNI}}}& \multirow{2}{*}{Abmil} & 47.2 & 51.5 & 67.0 & 68.4 & 60.1 & 65.3 & 93.7 & 93.2 & 96.6 & 97.4 \\
&  & ($\pm 3.0$) & ($\pm 5.0$) & ($\pm 6.0$) & ($\pm 4.0$) & ($\pm 5.0$) & ($\pm 8.0$) & ($\pm 2.0$) & ($\pm 2.0$) & ($\pm 2.0$) & ($\pm 1.0$) \\[3pt]
& \multirow{2}{*}{ClamMb} & 53.8 & 54.6 & 62.3 & 64.1 & 59.5 & 63.2 & 93.0 & 93.2 & 96.3 & 96.6 \\
&  & ($\pm 4.0$) & ($\pm 4.0$) & ($\pm 4.0$) & ($\pm 6.0$) & ($\pm 4.0$) & ($\pm 9.0$) & ($\pm 2.0$) & ($\pm 1.0$) & ($\pm 2.0$) & ($\pm 1.0$) \\[3pt]
& \multirow{2}{*}{ClamSb} & 51.9 & 56.2 & 66.9 & 65.1 & 57.9 & 58.2 & 92.2 & 93.2 & 96.9 & 97.9 \\
&  & ($\pm 6.0$) & ($\pm 6.0$) & ($\pm 4.0$) & ($\pm 5.0$) & ($\pm 6.0$) & ($\pm 3.0$) & ($\pm 2.0$) & ($\pm 2.0$) & ($\pm 2.0$) & ($\pm 1.0$) \\[3pt]
& \multirow{2}{*}{DSMIL} & 50.1 & 56.7 & 61.5 & 68.8 & 53.0 & 65.3 & 92.4 & 92.2 & 96.8 & 97.0 \\
&  & ($\pm 3.0$) & ($\pm 2.0$) & ($\pm 6.0$) & ($\pm 3.0$) & ($\pm 5.0$) & ($\pm 7.0$) & ($\pm 2.0$) & ($\pm 2.0$) & ($\pm 2.0$) & ($\pm 1.0$) \\[3pt]
& \multirow{2}{*}{TransMIL} & 52.6 & 58.0 & 59.1 & 63.4 & 66.0 & 62.2 & 92.9 & 93.0 & 94.6 & 96.5 \\
&  & ($\pm 3.0$) & ($\pm 5.0$) & ($\pm 4.0$) & ($\pm 4.0$) & ($\pm 5.0$) & ($\pm 5.0$) & ($\pm 2.0$) & ($\pm 2.0$) & ($\pm 2.0$) & ($\pm 2.0$) \\[3pt]
& \textbf{Mean} & 51.1 & \textbf{55.4} & 63.4 & \textbf{66.0} & 59.3 & \textbf{62.8} & 92.8 & \textbf{93.0} & 96.2 & \textbf{97.1} \\
\midrule
\multirow{12}{*}{\rotatebox[origin=c]{90}{\textbf{CONCH}}}& \multirow{2}{*}{Abmil} & 53.2 & 63.2 & 71.5 & 69.9 & 61.6 & 70.0 & 93.4 & 93.5 & 98.0 & 98.7 \\
&  & ($\pm 2.0$) & ($\pm 5.0$) & ($\pm 5.0$) & ($\pm 1.0$) & ($\pm 4.0$) & ($\pm 8.0$) & ($\pm 2.0$) & ($\pm 2.0$) & ($\pm 2.0$) & ($\pm 1.0$) \\[3pt]
& \multirow{2}{*}{ClamMb} & 59.2 & 63.2 & 69.1 & 71.2 & 62.6 & 69.3 & 93.4 & 93.6 & 98.7 & 98.7 \\
&  & ($\pm 6.0$) & ($\pm 5.0$) & ($\pm 2.0$) & ($\pm 4.0$) & ($\pm 8.0$) & ($\pm 5.0$) & ($\pm 1.0$) & ($\pm 2.0$) & ($\pm 1.0$) & ($\pm 1.0$) \\[3pt]
& \multirow{2}{*}{ClamSb} & 53.4 & 65.9 & 61.9 & 70.2 & 58.3 & 68.4 & 93.7 & 93.7 & 98.0 & 98.6 \\
&  & ($\pm 3.0$) & ($\pm 6.0$) & ($\pm 3.0$) & ($\pm 5.0$) & ($\pm 9.0$) & ($\pm 10.0$) & ($\pm 1.0$) & ($\pm 1.0$) & ($\pm 1.0$) & ($\pm 1.0$) \\[3pt]
& \multirow{2}{*}{DSMIL} & 57.7 & 61.9 & 68.9 & 74.2 & 61.7 & 68.6 & 91.7 & 93.3 & 98.3 & 98.2 \\
&  & ($\pm 4.0$) & ($\pm 6.0$) & ($\pm 6.0$) & ($\pm 2.0$) & ($\pm 7.0$) & ($\pm 5.0$) & ($\pm 1.0$) & ($\pm 2.0$) & ($\pm 1.0$) & ($\pm 1.0$) \\[3pt]
& \multirow{2}{*}{TransMIL} & 64.6 & 66.3 & 68.2 & 68.9 & 58.5 & 59.4 & 91.3 & 91.9 & 98.1 & 96.9 \\
&  & ($\pm 3.0$) & ($\pm 3.0$) & ($\pm 4.0$) & ($\pm 3.0$) & ($\pm 8.0$) & ($\pm 11.0$) & ($\pm 2.0$) & ($\pm 2.0$) & ($\pm 1.0$) & ($\pm 3.0$) \\[3pt]
& \textbf{Mean} & 55.9 & \textbf{64.1} & 67.9 & \textbf{70.9} & 60.5 & \textbf{67.1} & 92.7 & \textbf{93.2} & \textbf{98.2} & \textbf{98.2} \\

\bottomrule
\end{tabular}
\end{adjustbox}
\end{table*}

\end{document}